\documentclass[letterpaper]{article} 
\usepackage{aaai2026}  
\usepackage{times}  
\usepackage{helvet}  
\usepackage{courier}  
\usepackage[hyphens]{url}  
\usepackage{graphicx} 
\usepackage{natbib}  
\usepackage{caption} 
\usepackage{algorithm}
\usepackage{algorithmic}
\usepackage{amsfonts} 
\usepackage{booktabs}
\usepackage{subfigure}
\usepackage{amsfonts,amssymb}
\usepackage{amsmath}
\usepackage{graphicx} 
\newtheorem{theorem}{Theorem}
\newtheorem{asmp}{Assumption}
\usepackage{enumitem}
\usepackage{booktabs}
\usepackage{cleveref}
\usepackage{multirow} 
\usepackage{array}

\usepackage{newfloat}
\usepackage{listings}
\DeclareCaptionStyle{ruled}{labelfont=normalfont,labelsep=colon,strut=off} 
\floatstyle{ruled}
\newfloat{listing}{tb}{lst}{}
\floatname{listing}{Listing}
\title{Causality-inspired Federated Learning for Dynamic Spatio-Temporal Graphs}
\author {
  Yuxuan Liu\textsuperscript{\rm 1},
  Wenchao Xu\textsuperscript{\rm 2},
  Haozhao Wang\textsuperscript{\rm 3}\thanks{Corresponding author},
  Zhiming He\textsuperscript{\rm 1},
  Zhaofeng Shi\textsuperscript{\rm 1},\\
  Chongyang Xu\textsuperscript{\rm 1},
  Peichao Wang\textsuperscript{\rm 1},
  Boyuan Zhang\textsuperscript{\rm 1}
}
\affiliations {
  \textsuperscript{\rm 1} 
School of Information and Communication Engineering, University of Electronic Science and Technology of China, China\\
  \textsuperscript{\rm 2} Division of Integrative Systems and Design, Hong Kong University of Science and Technology, China\\
  \textsuperscript{\rm 3} School of Computer Science and Technology, Huazhong University of Science and Technology, China\\
  \{eie.yuxuan.liu, zfshi, cyxu, peichaowang, boyuanzhang\}@std.uestc.edu.cn, wenchaoxu@ust.hk, \\
  hz\_wang@hust.edu.cn,
  zmhe@uestc.edu.cn

}

\begin{document}

\maketitle

\begin{abstract}

  Federated Graph Learning (FGL) has emerged as a powerful paradigm for decentralized training of graph neural networks while preserving data privacy. However, existing FGL methods are predominantly designed for static graphs and rely on parameter averaging or distribution alignment, which implicitly assume that all features are equally transferable across clients, overlooking both the spatial and temporal heterogeneity and the presence of client-specific knowledge in real-world graphs. In this work, we identify that such assumptions create a vicious cycle of spurious representation entanglement, client-specific interference, and negative transfer, degrading generalization performance in Federated Learning over Dynamic Spatio-Temporal Graphs (FSTG). To address this issue, we propose a novel causality-inspired framework named SC-FSGL, which explicitly decouples transferable causal knowledge from client-specific noise through representation-level interventions. Specifically, we introduce a Conditional Separation Module that simulates soft interventions through client conditioned masks, enabling the disentanglement of invariant spatio-temporal causal factors from spurious signals and mitigating representation entanglement caused by client heterogeneity. In addition, we propose a Causal Codebook that clusters causal prototypes and aligns local representations via contrastive learning, promoting cross-client consistency and facilitating knowledge sharing across diverse spatio-temporal patterns. Experiments on five diverse heterogeneity Spatio-Temporal Graph (STG) datasets show that SC-FSGL outperforms state-of-the-art methods.

\end{abstract}


\section{Introduction}
Spatio-Temporal Graphs (STGs) model dynamic systems by jointly capturing spatial dependencies and temporal patterns. They are widely used in traffic forecasting, sensor networks, and mobility analytics~\cite{liang2023airformer,yao2019revisiting,lyu2025inco}. In practice, STG data are often distributed across regions or institutions, where data privacy concerns restrict centralized collection~\cite{FLSurveyandBenchmarkforGenRobFair_TPAMI24,Wang2024FedCDA,meng2024improving,qi2025cross}. This motivates the need for Federated Learning over Dynamic Spatio-Temporal Graphs (FSTGs), which enables decentralized training across clients without exposing raw data~\cite{s25072240,Wang2024FedCDA,FCCLPlus_TPAMI23,qi2023cross}. To ensure privacy preservation, Federated Graph Learning (FGL) offers a decentralized framework that trains Graph Neural Networks across clients while keeping local data private~\cite{MultiOrder_NeurIPS25,OASIS_NeurIPS25,LIU2025115106}.

\begin{figure}[t]

 \label{fig.background}
 \subfigure[Spatial heterogeneity]{
 
 \begin{minipage}[t]{0.47\columnwidth}
 \includegraphics[width=1\columnwidth]{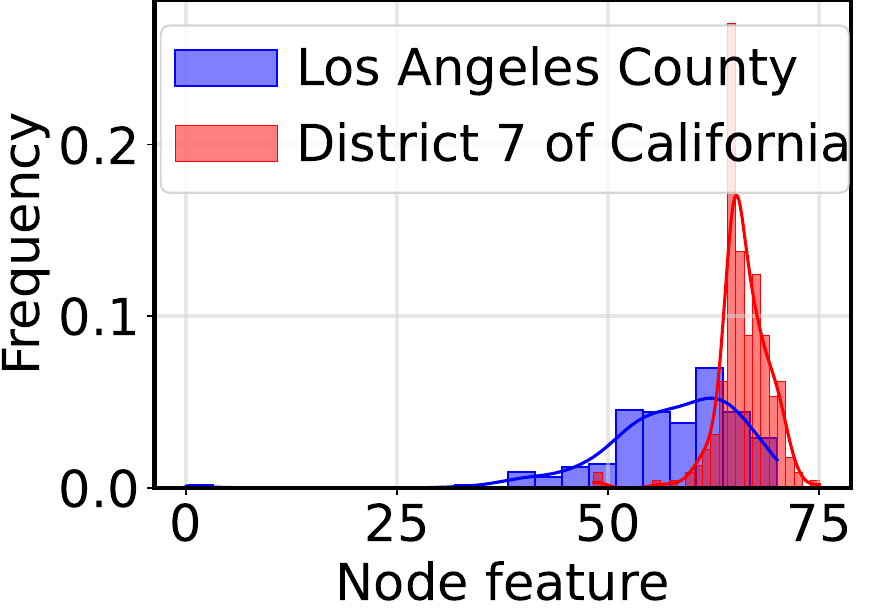}
 \end{minipage}%
 }
 \subfigure[Temporal heterogeneity]{
 \begin{minipage}[t]{0.47\columnwidth}
 \includegraphics[width=1\columnwidth]{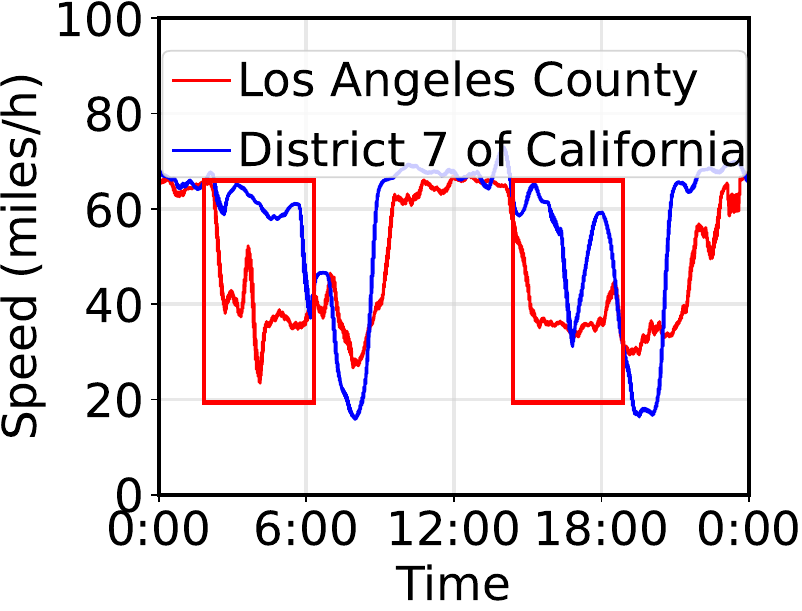}
 \end{minipage}%
 }
 \subfigure[Spatial causal features]{
 
 \begin{minipage}[t]{0.43\columnwidth}
 \includegraphics[width=1\columnwidth]{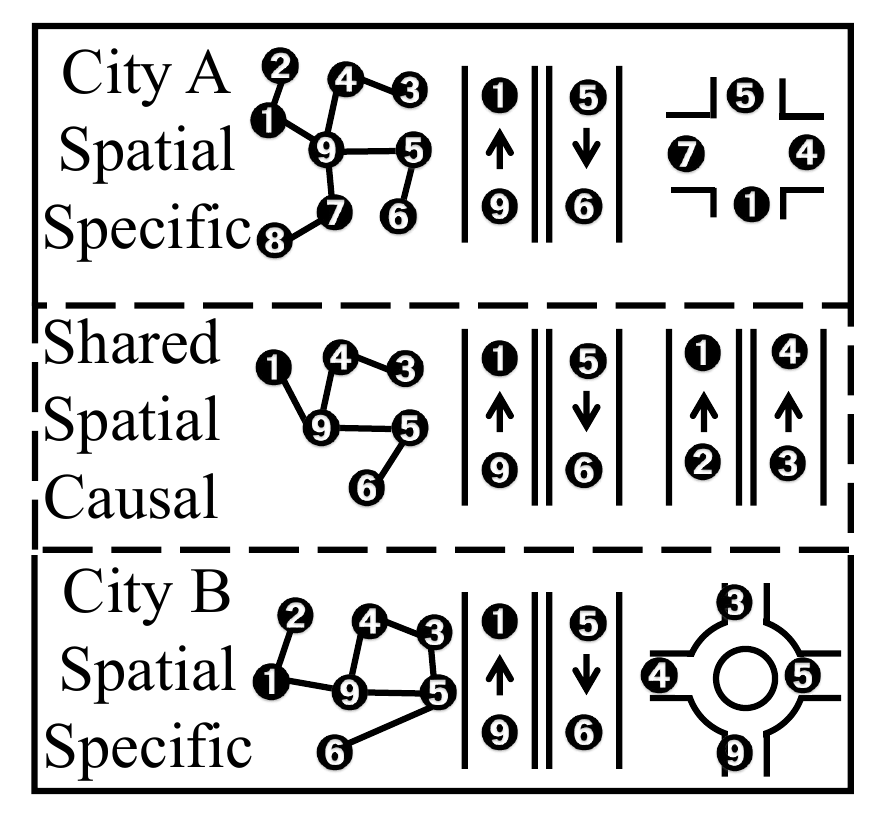}
 \end{minipage}%
 }
 \subfigure[Temporal causal features]{
 \begin{minipage}[t]{0.43\columnwidth}
 \includegraphics[width=1\columnwidth]{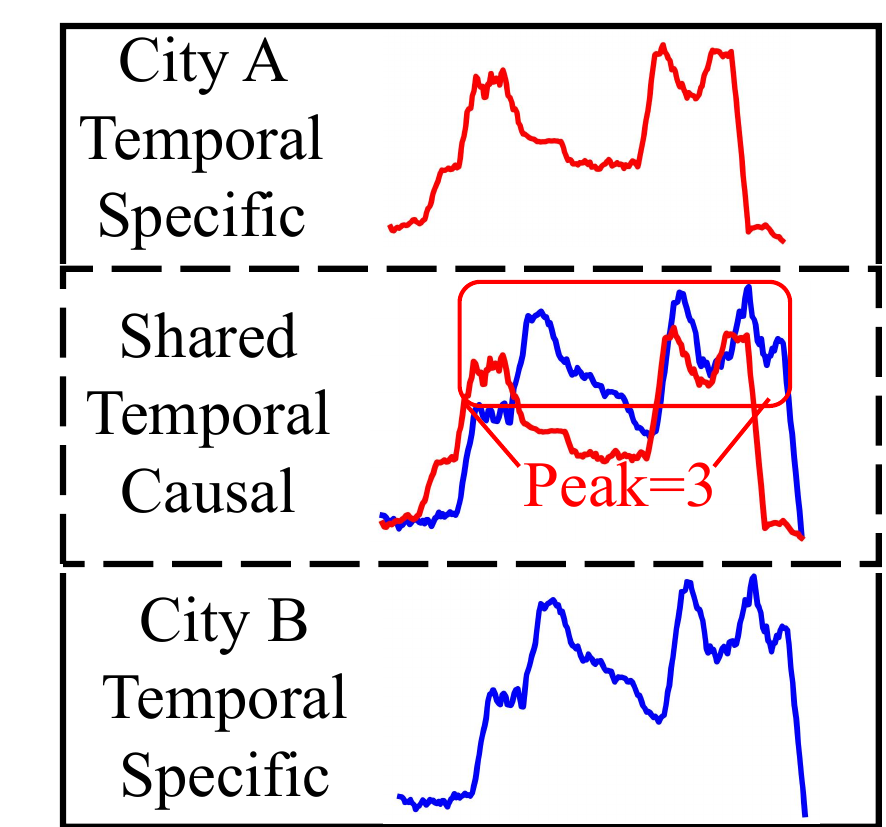}
 \end{minipage}%
 }
 \centering

 \caption{
Spatial and temporal heterogeneity across clients and shared causal patterns. \textbf{(a)} differences in graph structures (nodes and edges). \textbf{(b)} variation in traffic trends at the same time. \textbf{(c)} similar road layouts imply shared spatial causal structures. \textbf{(d)} recurrent temporal patterns suggest shared temporal causality.}

 \label{fig:background}
\end{figure}
Existing FGL can be broadly categorized into two paradigms: Graph-FL, where each client holds multiple individual graphs, and Subgraph-FL, where clients own local subgraphs of a global graph~\cite{openfgl_2024}. Across both paradigms, various strategies have been proposed to address statistical heterogeneity, including structural alignment~\cite{tan2023federated,NEURIPS2021_9c6947bd}, pseudo graphs~\cite{baek2023personalized,kim2025subgraphfederatedlearninglocal}, or adaptive client modules~\cite{AdaFGL_ICDE,FedTAD}. However, these techniques are primarily designed for static graphs and fall short when applied to dynamic environments. To support real-world systems where graphs evolve over time, recent works have extended FGL to FSTGs, incorporating temporal dynamics during federated training. For example, FUELS~\cite{liu2024personalized} adopts contrastive learning with dual semantic alignment to address distribution shifts across FSTGs, while FedCroST~\cite{Zhang_2024_TMC} introduces learnable spatio-temporal prompts to explicitly capture local STGs states. Despite such extensions, existing methods still rely on an assumption: \textit{client knowledge is equally transferable and semantically aligned.}

However, such assumptions fail to hold in FSTGs, which are inherently characterized by complex and non-stationary heterogeneity across clients. As illustrated in Figures~\ref{fig:background}a–b, different clients possess varied spatial structures and temporal dynamics, violating the assumed cross-client homogeneous alignment in both feature and distributional space. These mismatches trigger a vicious cycle during federated optimization: (1) \textit{Representation entanglement}, where generalizable patterns and specific variations are fused together in the learned features, making it hard to extract transferable knowledge; (2) \textit{Specific interference}, where the global model overfits to non-transferable local signals, introducing noise into other clients' updates; and (3) \textit{Negative transfer}, where improperly aggregated representations further degrade both global generalization and local adaptation. Such a self-reinforcing loop, where the inaccurate global model misguides local training and leads to even noisier representations and degraded aggregation quality, ultimately results in brittle personalization and poor performance. To address this fundamental challenge, we draw inspiration from recent advances in causal reasoning for graph learning~\cite{Lin_2022_CVPR,Lin_2021_ICML,zhao2024causalityinspired}, which emphasize disentangling stable causal mechanisms from spurious correlations. Building on these insights, we propose a novel categorization of causal knowledge in FSTGs: (1) \textit{Shared causal knowledge}, which refers to invariant spatio-temporal patterns that are stable across clients, e.g., recurring traffic peaks or universal road layouts; (2) \textit{Client-specific causal knowledge}, which captures factors unique to a particular region or system, e.g., city-specific road features (Figures~\ref{fig:background}c–d). Unfortunately, existing FGL methods ignore this distinction, treating all client knowledge as equally transferable, which leads to information leakage and false generalization. This raises a central question: \textit{How can we extract shared causal knowledge while suppressing client-specific patterns in heterogeneous FSTGs, in order to enhance generalization and reduce negative transfer?}

Based on the above motivations, we propose \underline{\textbf{S}}hared \underline{\textbf{C}}ausality-inspired \underline{\textbf{F}}ederated \underline{\textbf{S}}patio-Temporal \underline{\textbf{G}}raph \underline{\textbf{L}}earning (\textbf{SC-FSGL}), a novel framework that disentangles transferable and client-specific knowledge in FSTG learning without explicitly constructing structural causal models. Specifically, a Conditional Separation Module with a learnable soft mask adaptively extracts invariant and client-specific causal features, mitigating entangled representations and spurious correlations due to heterogeneity. Invariant features are globally aggregated to enhance generalization, while client-specific knowledge remain local to prevent negative transfer. To further promote global consistency, we introduce a causal codebook that aligns shared spatio-temporal causal embeddings via a contrastive objective, reducing cross-client inconsistency and fostering a unified semantic space. Our key contributions are:

\begin{itemize}
  \item We propose a causality-inspired representation learning framework for FSTGs, which adaptively separates shared and client-specific causal knowledge.
  
  \item We design a contrastive causal codebook to align spatial and temporal causal variables across clients, improving generalization and minimizing negative transfer under heterogeneous conditions.  

  \item We conduct extensive experiments on five real-world spatio-temporal datasets under statistically heterogeneous settings. SC-FSGL consistently outperforms state-of-the-art baselines across multiple metrics.
\end{itemize}

\section{Related Work}
\subsection{Causal Representation Learning in GNNs}
Causal inference improves generalization by identifying invariant mechanisms across environments. In graph learning, methods like OrphicX~\cite{Lin_2022_CVPR} and Gem~\cite{Lin_2021_ICML} disentangle causal and spurious relations in static graphs to enhance robustness. Extensions to STGs, such as DyGNNExplainer~\cite{zhao2024causalityinspired}, decompose dependencies into static and dynamic components but assume centralized training, overlooking FSTGs heterogeneity. Moreover, existing approaches often ignore the distinction between shared and client-specific causal patterns, leading to causal interference when transferring non-generalizable knowledge. To address this, we separate spatio-temporal causal features via conditional masking and contrastive alignment, supporting robust and personalized FSTG learning.\nocite{2025FedMFD}

\subsection{Spatio-temporal Graph Learning}
STG learning has been widely applied to traffic and mobility forecasting~\cite{zhang2025drawing,zhang2025coflownet}. Early models like DCRNN~\cite{li2017diffusion} and STGCN~\cite{yu2017spatio} integrate GCNs with recurrent or convolutional units to model dynamic dependencies. Later works such as GMAN~\cite{zheng2020gman}, \nocite{2025FedNCN,2025FedGCN}MegaCRN~\cite{jiang2023spatio}, STEAM~\cite{Gao_Xu_Gao_Cai_Ge_2025} and TWIST~\cite{wang2025twist} employ attention or meta-graph learning to enhance performance. However, most assume centralized or homogeneous data, lacking robustness to distribution shifts. Unlike them, we introduce a causal representation framework for heterogeneous FSTGs.

\subsection{Federated Graph Learning on STGs}
FGL is challenged by heterogeneous graph data across clients. Prior works address this via parameter reweighting~\cite{jiang2023spatio,FLSurveyandBenchmarkforGenRobFair_TPAMI24} or subgraph generation~\cite{zhang2021subgraph,baek2023personalized}, but often overlook structural invariants. FedStar~\cite{tan2023federated} extracts such invariants in static graphs, yet fails to handle dynamic STG settings. Recent approaches~\cite{yuan2022fedstn,Zhang_2024_TMC,liu2024personalized} model local spatio-temporal distributions via global networks, prompts or contrastive learning. However, they focus on distributional alignment without explicitly modeling transferable causal patterns. Our method instead disentangles shared and client-specific causal representations, enabling more robust generalization under spatio-temporal heterogeneity.

\section{Problem Description}

\subsection{Problem setting \& notations}
At time step $t$, the input STG $\mathbf{\mathcal{G}}^k_t=\{X^k_t,A^k_t\}$ of client $k$ ($k \in K$) comprises a historical node feature matrix $\mathbf{X}^k_t\in \mathbb R^{\vert V^k\vert\times d}$ and an adjacency matrix $\mathbf{A}^k_t \in \mathbb R^{\vert V^k\vert \times \vert V^k\vert}$, where $K$ represents the total number of clients, $V^k$ represents the sets of nodes feature and $d$ represents the dimension of node features. The prediction results at time $t$, denoted by $\mathbf{\mathcal{Y}}_t^k=\{{\hat X^k_t,A^k_t }\}$, includes predicted future node features $\mathbf{\hat X}^k_t\in \mathbb R^{\vert V^k\vert\times d}$. At this point, our local task is to train a model $f_{\theta^k}$ with model parameters $\theta^k$. This model aims to establish causal relationships between variables based on the historical $\gamma$ steps to predict the future $\beta$ steps of spatio-temporal states:
\begin{equation}
 \{{X^k_{t-\gamma},X^k_{t-\gamma+1},...,X^k_t}\}\mathop{\rightarrow}\limits_{\theta^k}^{f_{\theta^k}} \{{\hat X^k_{t+1},\hat X^k_{t+2},...,\hat X^k_{t+\beta}}\},
\end{equation}

\subsection{Causal View of FSTGs}\label{sec:causal_view}
\begin{figure}[t]
 \centering
 \includegraphics[width=1\linewidth]{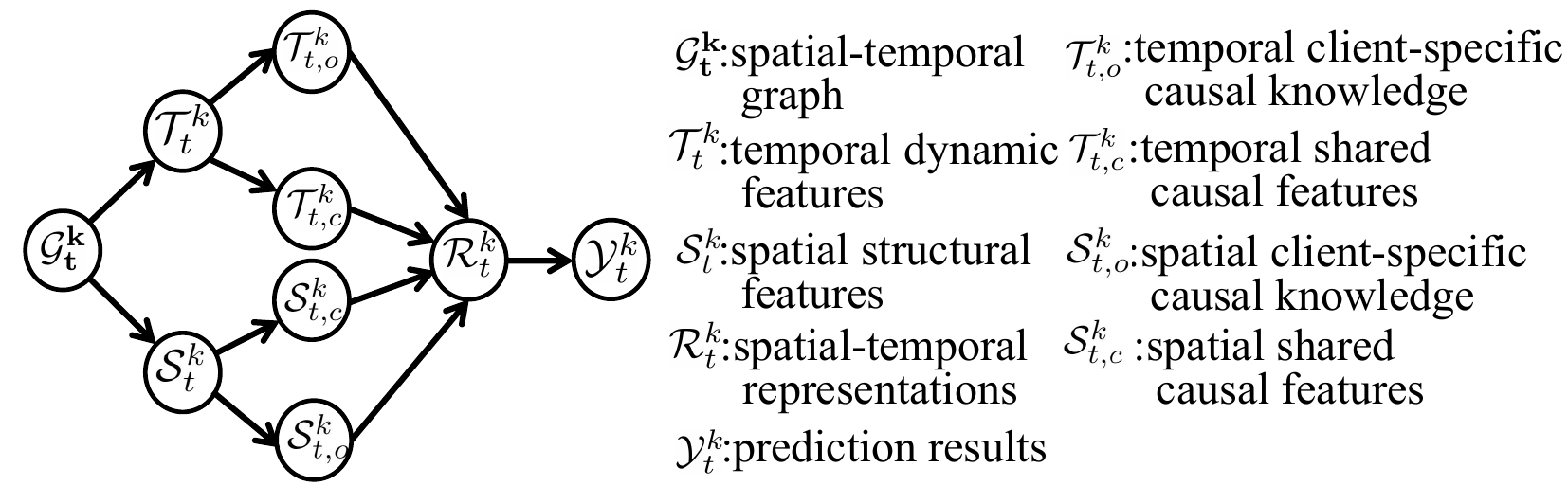}

 \caption{
 A conceptual illustration inspired by the SCM on client k, showing the relationships among observed variables and latent components.
 }
 \label{fig:Causal Grpah}

\end{figure}

In FSTGs, each client $k~(k\in K)$ is treated as an environment with distinct data-generating processes (See Figure~\ref{fig:Causal Grpah}), and $K$ is the set of participating clients. We decompose the spatial and temporal variables $\mathcal{S}_t^k$, $\mathcal{T}_t^k$ into shared and client-specific components:
\begin{equation}
  \mathcal{S}_t^k = \mathcal{S}_{t,c}^k \cup \mathcal{S}_{t,o}^k, \quad \mathcal{T}_t^k = \mathcal{T}_{t,c}^k \cup \mathcal{T}_{t,o}^k.
\end{equation}
To analyze the effect of shared causal variables, we adopt the Structural Causal Model (SCM) framework~\cite{pearl2009causality}. In this framework, the \textbf{do}-operator $\textbf{do}(\cdot)$ denotes an intervention that sets a variable to a fixed value and removes all incoming causal influences. Using this formalism, the interventional distribution $P(\mathcal{Y}_t^k \mid \textbf{do}(\mathcal{T}_{t,c}^k))$ quantifies the causal effect of the shared temporal variables on the outcome. Under the assumption that $\mathcal{T}_{t,o}^k$ and $\mathcal{S}_t^k$ are conditionally independent given $\mathcal{G}^k_{t}$, we approximate the interventional distribution via observational distributions as:
\begin{equation}
 \begin{split}
P(\mathcal{Y}_t^k |\textbf{do}(\mathcal{T}_{t,c}^k)) &= \sum P(\mathcal{Y}_t^k | \textbf{do}(\mathcal{T}_t^k)) P(\mathcal{T}_{t,c}^k)\\
      &= \sum P(\mathcal{T}_{t,o}^k) \sum P(\mathcal{T}_t^k | \mathcal{G}_t^k) P(\mathcal{S}^k_t),
\end{split}
\label{eq.adj_T}
\end{equation}
where $\mathcal{G}_t^k$ denotes the observed graph. Similarly, the interventional distribution with respect to the spatial is given by $P(\mathcal{Y}_t^k | \textbf{do}(\mathcal{S}_{t,c}^k)) = \sum P(\mathcal{T}_t^k) \sum P(\mathcal{Y}_t^k | \mathcal{G}_t^k) P(\mathcal{T}_t^k)$. See Appendix for details.

Since direct interventions are impractical in the federated scenarios, we introduce a learnable soft mask $M_t^k \in [0,1]^d$ to approximate intervention effects in latent space:
\begin{equation}
  \Phi(X_t^k) = (1 + \text{LN}(M_t^k)) \odot X_t^k,
\end{equation}
where $\text{LN}(\cdot)$ denotes layer normalization and $\odot$ is the Hadamard product. This soft mask simulates intervention by attenuating the influence of client-specific variables, thereby approximating $\textbf{do}(X_t^k = x_t^k)$ in representation space.

To promote the invariance of retained features across heterogeneous clients, we adopt the principle of \textit{Invariant Risk Minimization (IRM)}~\cite{arjovsky2020invariantriskminimization,pmlr-v119-chang20c}, which encourages a predictor $\omega$ to remain optimal across all environments. Formally, this principle seeks a feature representation $\Phi$ such that a single predictor achieves optimality under all environment-specific risks:
\begin{equation}
\label{eq:IRM1}
\min_{\Phi} \sum_{k \in K} \mathcal{R}_k(\omega , \Phi) \quad \text{s.t.} \quad \omega \in \arg\min_{\bar{\omega}} \mathcal{R}_k(\bar{\omega} , \Phi),
\end{equation}
where $R_k(\omega, \Phi) = \mathbb{E}_{(x,y) \sim \mathcal{D}_k}[\ell(\omega(\Phi(x)), y)]$ denotes the prediction risk of client $k$, with $\omega$ being a linear classifier and $\Phi$ the shared encoder. This constraint enforces that $\Phi$ should extract invariant features that enable $\omega$ to generalize across all client distributions.

\begin{figure*}[t]
\label{fig.framework of FedBook}
\centering
\includegraphics[width=1\linewidth]{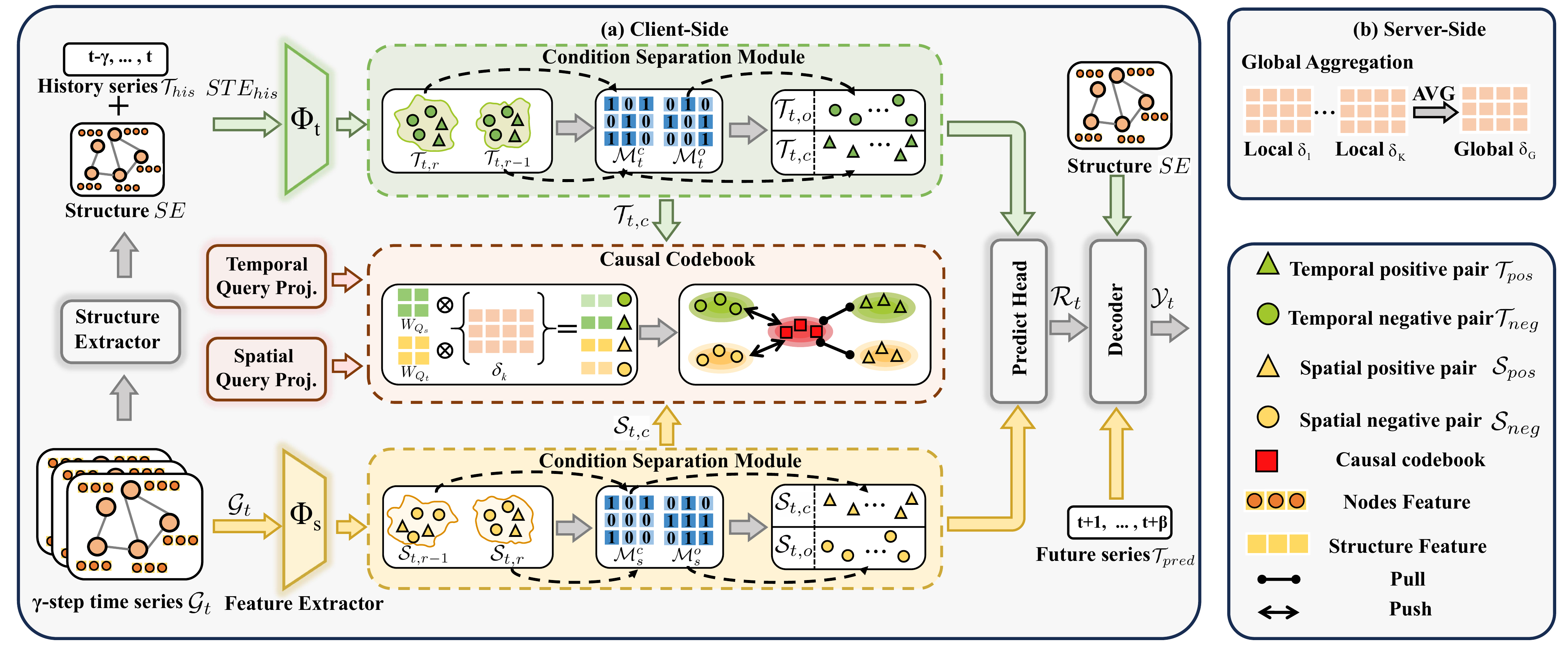}

\caption{The figure illustrates the overall architecture of SC-FSGL, including the client-side feature extraction, causal codebook construction, and global aggregation process. The model extracts spatio-temporal features, separates shared and client-specific causal variables, and leverages the causal codebook to enhance the prediction accuracy through soft intervention and contrastive representation alignment.}
\label{fig:Overview}

\end{figure*}

\section{Methodology}
In this section, we introduces the SC-FSGL in the Figure~\ref{fig:Overview}, a prediction model for FSTG that incorporates shared causal relationships to enhance accuracy and mitigate the impact of client-specific causal knowledge on the global model. Next, we will delineate the various modules of the SC-FSGL.

\subsection{Spatio-Temporal Embedding ($STE$)}
The spatial network structure and historical observations play a crucial role in FSTG prediction. To encode spatial information, we employ node2vec~\cite{grover2016node2vec} as the \textbf{Structure Extractor}, generating spatial embeddings ($SE$) that preserve the graph topology. For temporal representation, we encode historical timestamps into $\mathcal{T}^k_{his} \in \mathbb{R}^{\gamma \times \vert V^k\vert \times D}$ using a week-day-hour format, and similarly encode future time steps into $\mathcal{T}^k_{pred} \in \mathbb{R}^{\beta \times \vert V^k\vert \times D}$, where $\vert V^k\vert$ is the number of nodes and $D$ the embedding dimension. We then concatenate temporal and spatial embeddings to obtain time-varying vertex representations following GMAN~\cite{zheng2020gman}: $STE^k_{his} = \textbf{concat}[\mathcal{T}^k_{his}, SE^k]$ and $STE^k_{pred} = \textbf{concat}[\mathcal{T}^k_{pred}, SE^k]$, which are input into the Temporal Feature Extractor $\Phi^k_t$. Meanwhile, the $\mathcal{G}^k_t$ is processed by the Spatial Feature Extractor $\Phi^k_s$ to generate the spatial representation $\mathcal{S}^k_t$.

\subsection{Conditional Separation Module}

To effectively disentangle transferable causal patterns from client-specific spurious factors, we propose a Conditional Separation Module, based on the assumption that shared causal factors remain stable across communication rounds, while client-specific patterns tend to fluctuate due to local heterogeneity. However, in practice, these stable causal patterns are often entangled with noisy client-specific signals due to variations in data distributions and latent confounders. To address this, we introduce a \textit{soft intervention mechanism} that simulates causal interventions in the latent space, selectively attenuating spurious influences and facilitating the extraction of invariant causal representations.

 Specifically, the Conditional Separation Module comprises two symmetric submodules, applied respectively to temporal and spatial representations. Given an input feature $ \mathcal{F}_{t}^{k} \in \{\mathcal{T}_{t}^{k}, \mathcal{S}_{t}^{k}\} $, we denote $ \mathcal{F}_{t, r-1}^k $ as the representation obtained at the previous round $r{-}1$, serving as a reference to extract causal patterns that are invariant across rounds, where $r$ represents the current communication round. Then, we learn a soft mask $\mathcal{M}_{t}^{k,c}=\sigma(\mathcal{F}_{t,r-1}\omega_{\mathcal{F}}+b_{\mathcal{F}})$ that highlights shared causal dimensions, and a complementary mask $\mathcal{M}_{t}^{k,o} = 1 - \mathcal{M}_{t}^{k,c}$ to capture client-specific components. The $\omega_{\mathcal{F}}$ denotes the weight matrix and $b_{\mathcal{F}}$ the bias term of a MLP, $\sigma(\cdot)$ is a nonlinear activation function. These masks are applied to the input representations to disentangle causal and spurious components via:
\begin{equation}
\left\{ 
\begin{aligned}
\mathcal{F}_{t,c}^{k} &= \sigma\left[(1 + \mathrm{LN}(\mathcal{M}_{t}^{k,c})) \odot \mathcal{F}_{t,r}^{k} \right] \\
\mathcal{F}_{t,o}^{k} &= \sigma\left[(1 + \mathrm{LN}(\mathcal{M}_{t}^{k,o})) \odot \mathcal{F}_{t,r}^{k} \right],
\end{aligned}
\right.
\end{equation}
where $\mathrm{LN}(\cdot)$ is layer normalization. This decomposition enables the model to approximate causal interventions by suppressing client-specific noise, thus promoting the extraction of stable, transferable features across clients.

To guide the mask learning, we jointly optimize two objectives: 
(1) a \textit{contrastive alignment loss} that encourages the shared representations $\mathcal{F}_{t,c}^k$ to be semantically consistent across clients, using the global causal codebook $\delta$ as a reference (detailed in the next section); 
(2) an \textit{IRM-inspired regularization} that promotes representation invariance across heterogeneous client environments by penalizing the gradient of the prediction loss with respect to the predictor $\omega$. Following~\cite{arjovsky2020invariantriskminimization}, we relax the original IRM constraint into a differentiable surrogate objective using a gradient penalty, which encourages the predictor to remain optimal across different environments. Formally, the IRM loss is defined as:
\begin{equation}
  \mathcal{L}_{\text{irm}} = \sum_{\Phi_{\delta}\in \{\Phi_s,\Phi_t\}}\left\| \nabla_{\omega} \mathcal{L}_{\text{pred}}(\omega \circ \Phi_{\delta}) \right\|_2^2,
\end{equation}
where $\mathcal{L}_{\text{pred}}$ is the prediction loss, measured by MAE between predicted values and ground truth. $\Phi_\delta$ is the encoder aligned with the codebook $\delta$, and $\omega$ is the prediction layer. By minimizing this gradient norm, the model enforces stability in predictive performance across clients, thereby reinforcing the extraction of invariant causal features.

\subsection{Causal Codebook}\label{sec:codebook}
Existing methods often increase the amount of information fed into the model, leading the model to overly focus on client-specific knowledge that is not beneficial to the current client during the aggregation phase~\cite{qi2025federated,wangbsemifl,liu2024disentangle}. To reduce the interference of client-specific causal parts and enhance the contribution of shared causal knowledge, we propose a causal codebook $\delta \in \mathbb{R}^{ \phi \times d}$, where $\phi$ is the number of causal prototypes and $d$ is the feature dimension of items. This is a shared learnable prototype matrix that acts as a \textit{semantic anchor}, i.e., a set of latent reference vectors that define a consistent representation space to align spatial and temporal causal features across clients. To align causal variables with $\delta$, we compute similarity scores between projected features and codebook:
 \begin{equation}
 \label{eq:Querry}
\left\{ 
\begin{array}{ll} 
 \alpha^s_{j}=\frac{exp((W_{Q_s}*\mathcal{S}^k_{t,c}+b)*\delta_{j}^\top)}{\sum_{j=1}^{\phi}exp(((W_{Q_s}*\mathcal{S}^k_{t,c}+b)*\delta_{j}^\top)}\\
 \alpha^t_{j}=\frac{exp((W_{Q_t}*\mathcal{T}^k_{t,c}+b)*\delta_{j}^\top)}{\sum_{j=1}^{\phi}exp(((W_{Q_t}*\mathcal{T}^k_{t,c}+b)*\delta_{j}^\top)}, 
\end{array} 
\right.
\end{equation}
where $j$ denote the row indices of $\delta$, $W_Q$ represents the learnable parameters of the local Query model, which performs a projection transformation of $\mathcal{T}^k_{t,c}$ and $\mathcal{S}^k_{t,c}$. Considering the causal relationship between the spatial causal variables and the temporal causal variables and their impact on the final result $\mathcal{Y}^k_t$, we further strengthen the shared spatial and temporal causal relationships in local model. The top-1 ($max$) and second-best ($\hat{max}$) matching entries in the codebook are used as positive and negative pairs:
 \begin{equation}
 \label{eq:pos_neg}
\left\{ 
\begin{array}{ll} 
 \mathcal{S}^k_{pos}=max_{j}(\alpha^s_{j}*\delta_{j})\\
 \mathcal{S}^k_{neg}=\hat{max}_j(\alpha^s_{j}*\delta_{j}) 
\end{array} 
\right.
\left\{ 
\begin{array}{ll} 
 \mathcal{T}^k_{pos}=max_j(\alpha^t_{j}*\delta_{j})\\
 \mathcal{T}^k_{neg}=\hat {max}_j(\alpha^t_{j}*\delta_{j}) .
\end{array} 
\right.
\end{equation}

Based on the selected pairs, we define a contrastive loss to align $\mathcal{S}^k_{t,c}$ and $\mathcal{T}^k_{t,c}$ with their positive codebook entries while separating them from negatives:
\begin{equation}
\small
\label{eq:L_com}
 \begin{split}
\mathcal{L}_{\text{com}}& = \log \frac{\exp(\mathrm{sim}(\delta, \mathcal{S}_{\text{pos}}^k)/\tau)}{\exp(\mathrm{sim}(\delta, \mathcal{S}_{\text{pos}}^k)/\tau) + \sum \exp(\mathrm{sim}(\delta, \mathcal{S}_{\text{neg}}^k)/\tau)}  \\
&+\log \frac{\exp(\mathrm{sim}(\delta, \mathcal{T}_{\text{pos}}^k)/\tau)}{\exp(\mathrm{sim}(\delta, \mathcal{T}_{\text{pos}}^k)/\tau) + \sum \exp(\mathrm{sim}(\delta, \mathcal{T}_{\text{neg}}^k)/\tau)} ,
 \end{split}
\end{equation}
where $\mathrm{sim}(\cdot)$ denotes cosine similarity and $\tau$ is the temperature parameter. For each client, the top-1 matched prototype serves as the positive anchor, while the second-best is used as a hard negative, following hard negative mining to enhance discriminability. This encourages shared causal features to align with semantically consistent prototypes while being separated from close but suboptimal ones.

\subsection{Local Training and Inference}
\label{section:Local Training}
During the training phase, the client’s encoder represents $\mathcal{R}^k_t$ as input to the decoder for predicting future value $\mathcal{Y}^k_t$. To minimize the discrepancy between the predicted output and the actual ground truth, we define a predictor loss function as $ \mathcal{L}_{mse}={\vert\vert \mathcal{R}^k_t-\mathcal{Y}^k_t\vert\vert^2}$. Finally, we formulate the complete local loss function, which integrates additional components for comprehensive error minimization:
\begin{equation}
\mathcal{L}_{\text{local}} = \mathcal{L}_{\text{mse}} + \alpha  \mathcal{L}_{\text{com}} + \beta \mathcal{L}_{irm} ,
\label{eq:final_objective}
\end{equation}
where $\mathcal{L}_{\text{com}}$ aligns shared causal representations via contrastive learning, and the IRM penalty enforces predictor invariance across clients. The $\alpha$ and $\beta$ are hyperparameters that balance the influence of the contrastive loss and the IRM regularization term, respectively. During local training, clients extract spatio-temporal features, apply conditional separation to obtain disentangled representations, and perform prediction. The codebook $\delta$ is updated via federated averaging and used to align causal features during local training. 

\begin{table*}[t]
\centering
\small
\setlength{\tabcolsep}{4pt} 
\begin{tabular}{l|cc|cc|cc|cc|cc|cc}
\toprule
\multirow{1}{*}{\textbf{Client}} 
& \multicolumn{2}{c|}{\textbf{METRLA}} 
& \multicolumn{2}{c|}{\textbf{PEMSD4}} 
& \multicolumn{2}{c|}{\textbf{PEMSD7(M)}} 
& \multicolumn{2}{c|}{\textbf{PEMSD8}} 
& \multicolumn{2}{c|}{\textbf{PEMSBAY}} 
& \multicolumn{2}{c}{\textbf{Avg.}} \\
\midrule
\textbf{Metric} & \textbf{MAE} & \textbf{MAPE} & \textbf{MAE} & \textbf{MAPE} & \textbf{MAE} & \textbf{MAPE} & \textbf{MAE} & \textbf{MAPE} & \textbf{MAE} & \textbf{MAPE} & \textbf{MAE} & \textbf{MAPE} \\
\midrule
FedAvg   & 12.21 & 23.58\% & 3.41 & 7.13\% & 5.02 & 14.25\% & 3.72 & 7.22\% & 3.44 & 7.69\% & 5.56 & 11.97\% \\
FedProx  & 7.80 & 19.37\% & 3.20 & 6.83\% & 7.15 & 16.22\% & 3.09 & 6.70\% & 4.35 & 10.15\% & 5.12 & 11.85\% \\
Moon    & 7.54 & 18.00\% & 2.65 & 5.57\% & 6.21 & 15.66\% & 3.11 & 5.84\% & 4.35 & 8.11\% & 4.77 & 10.64\% \\
FedRep   & 6.78 & 17.71\% & 8.88 & 14.95\% & 4.37 & 12.15\% & 2.46 & 5.79\% & 7.89 & 13.67\% & 6.07 & 12.85\% \\
FedStar  & 28.71 & 56.69\% & 4.35 & 8.17\% & 16.41 & 34.24\% & 3.25 & 7.16\% & 33.71 & 53.31\% & 17.29 & 31.91\% \\
GMAN    & 9.51 & 22.26\% & 3.62 & 6.60\% & 3.64 & 10.53\% & 3.12 & 5.84\% & 3.29 & 6.45\% & 4.64 & 10.34\% \\
MegaCRN  & 10.39 & 35.54\% & 4.39 & 10.75\% & 8.84 & 29.52\% & 3.78 & 8.54\% & 5.10 & 14.05\% & 6.50 & 19.68\% \\
FUELS   & 7.30 & 18.84\% & 4.87 & 8.44\% & 4.70 & 12.12\% & 3.36 & 6.56\% & 2.34 & 4.92\% & 4.51 & 10.18\% \\
\midrule
Local   & 7.08 & \textbf{17.46\%} & 4.17 & 9.75\% & 7.11 & 14.81\% & 2.54 & 6.06\% & \textbf{2.39} & \textbf{6.13\%} & 4.66 & 10.84\% \\
\textbf{SC-FSGL} & \textbf{6.38} & 17.57\% & \textbf{2.46} & \textbf{4.65\%} & \textbf{3.50} & \textbf{9.94\%} & \textbf{2.29} & \textbf{5.72\%} & 6.44 & 12.53\% & \textbf{4.21} & \textbf{10.08\%} \\
\bottomrule
\end{tabular}
\caption{Performance comparison of different approaches for FSTG on various datasets~(Clients).}
\label{tab:client-performance}
\end{table*}

\subsection{Convergence Analysis}
In this section, we analyze and prove the convergence of the model. We have the following assumptions: 
\begin{asmp}
\label{assum.1}
The objective function $L_{k}$ is convex,
  \begin{align}\label{eq:theorem1}
     L_k(a) \geq L_k(b)+\langle\nabla L_k(b), a-b\rangle.
  \end{align}
\end{asmp}

\begin{asmp}
\label{assum.2}
For $L_k$, all gradients of the model parameter $\theta$ associated with it are constrained by a constant $M$,
  \begin{align}\label{eq:theorem2}
     \mathbb E(||\nabla L_{k}(\theta)||^2)\leq M^2.
  \end{align}
\end{asmp}
Suppose there are $K$ participating clients, $\mathcal{G}_{k,t}$ representing the STG input of client k($k\in K$) at the edge of time $t$, and the local model parameter is defined as $\theta^k_r=\{{\theta^k_{r,a},\theta_{r,b}}\}$, where $r\in R$ represents the aggregation Round. $\theta_{r, a}$ represents a local model parameter that does not participate in aggregation, $\theta_{r,b}$ is a model parameter of causal codebook, and $\theta_{r,b}$ participates in global aggregation. If $L_k$ is a convex function, then $\theta$ is a convex set, and we assume a bound between the model parameters and the optimal model parameters, where $||\theta_r^k-\theta^{k,*}||\leq I$. Partial loss function is $L (\theta_r ^ k) = L (\theta_ {r, a} ^ k) + L (\theta ^ k_ {r, b})$.
\begin{theorem}
causal codebook converges to the following bound when the learning rate is $\eta$, 
\label{Theorem:Gradient}
    \begin{align}
    \label{equation:Gradient}
    \frac{1}{R}\sum_{r=1}^{R}[L_k(\theta^k_{r,b})-L_k(\theta_{b}^{k,*})] \leq I^2 \frac{1}{2R\eta} + \frac{M^2}{2}\eta.
    \end{align}
\end{theorem}
Detailed proofs are provided in the Appendix.

\section{Experiments}
\subsection{Experimental settings}

\noindent\textbf{Datasets.} We introduce the datasets used in our experimentation, which comprise real traffic data from five distinct cities: METRLA~\cite{li2017diffusion}, PEMSD4~\cite{guo2019attention}, PEMSD7(M)~\cite{yu2017spatio}, PEMSD8~\cite{guo2019attention}, and PEMSBAY~\cite{li2017diffusion}. Each dataset corresponds to one client to preserve strong spatio-temporal heterogeneity. For details on dataset statistics, preprocessing, and partitioning protocols, please refer to the Appendix.

\noindent\textbf{Baseline.} We compared our approach with state-of-the-art methods, including the baseline FedAvg~\cite{mcmahan2017communication}, methods addressing federated data heterogeneity such as FedProx~\cite{li2020federated}, FedRep~\cite{collins2021exploiting}, Moon~\cite{li2021model}, and FedStar~\cite{tan2023federated}, as well as models for spatio-temporal prediction tasks like GMAN~\cite{zheng2020gman}, MegaCRN~\cite{jiang2023spatio} and FUELS~\cite{liu2024personalized}. 

\noindent\textbf{Metrics.} We evaluate SC-FSGL using three standard metrics: Mean Absolute Error (MAE), Root Mean Squared Error (RMSE), and Mean Absolute Percentage Error (MAPE).

\begin{table}[h]
\centering
\small 
\resizebox{\linewidth}{!}{ 
\begin{tabular}{l|cc|cc|cc}
\toprule
\multirow{1}{*}{\textbf{Time}} & \multicolumn{2}{c|}{\textbf{60 min}} & \multicolumn{2}{c|}{\textbf{30 min}} & \multicolumn{2}{c}{\textbf{15 min}} \\
\midrule
\textbf{Metric} & \textbf{MAE} & \textbf{MAPE} & \textbf{MAE} & \textbf{MAPE} & \textbf{MAE} & \textbf{MAPE} \\ 
\midrule
FedAvg  & 5.56 & 11.97\% & 4.80 & 10.99\% & 5.32 & 10.83\% \\
FedProx & 5.12 & 11.85\% & 4.68 & 9.85\% & 4.15 & 9.04\% \\
Moon   & 4.77 & 10.64\% & 5.12 & 10.95\% & 5.31 & 13.30\% \\
FedRep  & 6.07 & 12.85\% & 3.93 & 8.48\% & 4.55 & 10.83\% \\
FedStar & 17.29 & 31.91\% & 23.32 & 42.59\% & 19.93 & 35.00\% \\
GMAN   & 4.64 & 10.34\% & 3.92 & 8.80\% & 3.64 & 8.34\% \\
MegaCRN & 6.50 & 19.68\% & 6.38 & 19.55\% & 6.33 & 19.46\% \\
FUELS  & 4.51 & 10.18\% & 4.24 & 10.31\% & 3.87 & 8.80\% \\
\midrule
Local  & 4.66 & 10.84\% & 5.00 & 10.92\% & 4.64 & 13.09\% \\
\textbf{SC-FSGL} & \textbf{4.21} & \textbf{10.08\%} & \textbf{3.67} & \textbf{8.19\%} & \textbf{3.29} & \textbf{7.76\%} \\
\bottomrule
\end{tabular}
}
\caption{Performance across prediction horizons.}
\label{tab:overall-performance}

\end{table}
\noindent\textbf{Implementation Details.}
For all federated learning methods, communication rounds were set to early stop, and local training consisted of one epoch. Our implementation, developed using PyTorch, was executed on two NVIDIA 3090 GPUs for all experiments. Each dataset has unique traffic network structures and timestamps. 

\subsection{The Results of SC-FSGL}

\textbf{Performance Across Prediction Times.} To comprehensively evaluate SC-FSGL, we compare it with eight representative baselines across three prediction horizons 60, 30, and 15 minutes, on MAE and MAPE metrics (see Table~\ref{tab:overall-performance}). SC-FSGL consistently achieves the best performance across all metrics and time spans. For 60 minute forecasts, it has the lowest MAE (4.21), outperforming FUELS (4.51) and GMAN (4.64), as well as the MAPE (10.08\%). Similar trends are observed at 30 and 15 minutes, where SC-FSGL achieves MAE of 3.67 and 3.29, respectively, surpassing both spatio-temporal (e.g., GMAN) and federated (e.g., FedProx, Moon) baselines. Notably, methods like FedRep and Moon, though partially addressing heterogeneity, suffer from unstable performance due to lacking causal disentanglement. Models such as FedStar and MegaCRN underperform in all settings, indicating poor robustness under spatiotemporal heterogeneity. In contrast, SC-FSGL’s consistent superiority confirms the effectiveness of its causal disentanglement and alignment mechanisms in heterogeneous FSTG learning.\\
\textbf{The prediction performance on different clients.} To evaluate SC-FSGL’s robustness under spatio-temporal heterogeneity, we compare its predictive accuracy across five real-world traffic datasets, each representing a distinct client. As shown in Table~\ref{tab:client-performance}, SC-FSGL achieves the best or second-best results. For instance, on PEMSD4 and PEMSD8, SC-FSGL achieves the lowest MAEs of 2.46 and 2.29, outperforming both federated (e.g., FedAvg, FedProx) and spatio-temporal models (e.g., GMAN, FUELS). On PEMSD7(M), it achieves a MAE of 3.50, significantly better than Moon (6.21) and MegaCRN (8.84), demonstrating resilience to temporal drift. SC-FSGL effectively extracts shared causal features, mitigating negative transfer and enhancing generalization. When averaged across all clients, it attains the best overall score.

\begin{figure*}[t]
 \centering
 \subfigure[]{
 \begin{minipage}[t]{0.32\textwidth}
 \centering
 \includegraphics[width=\textwidth]{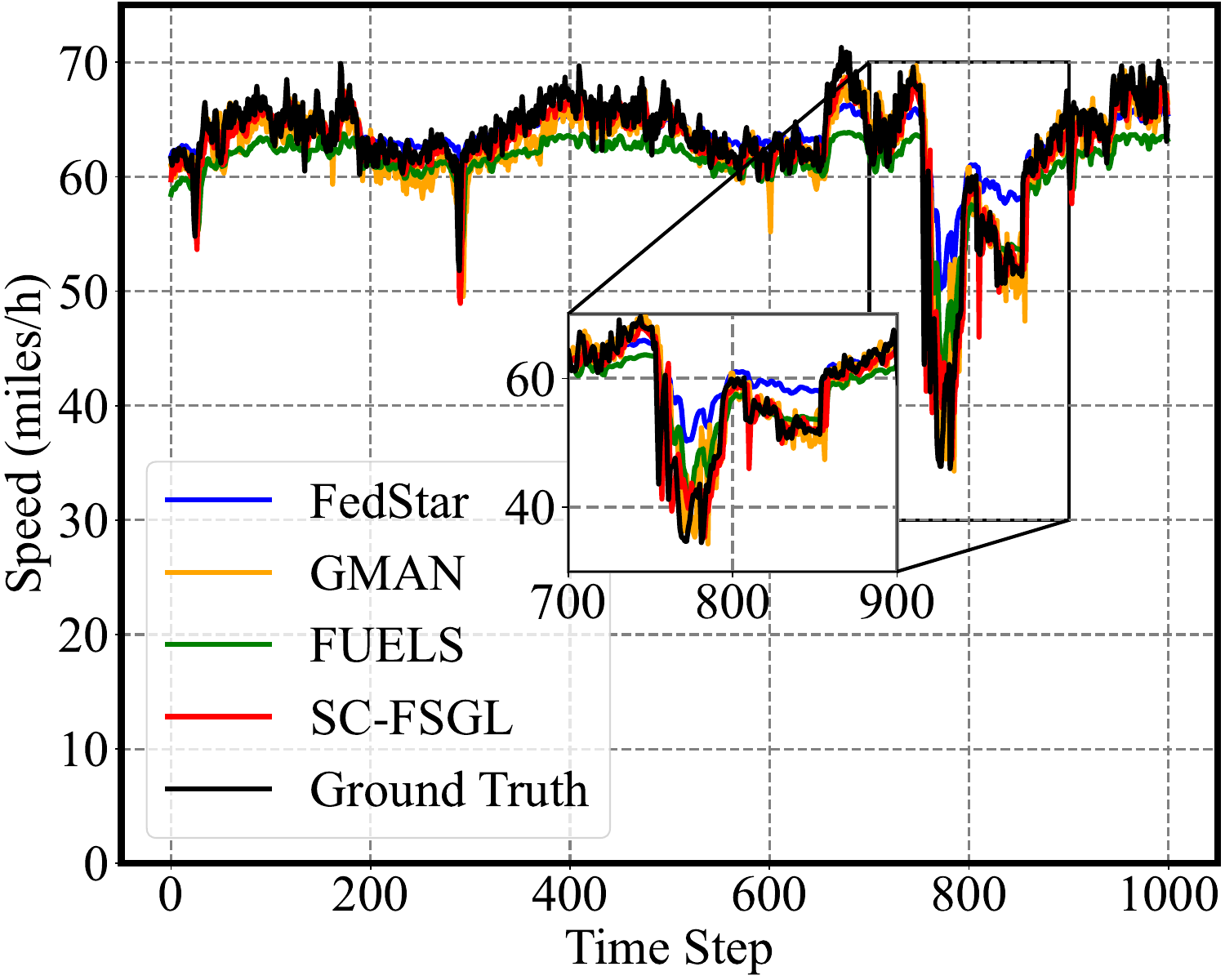}
 \end{minipage}
 }
 \subfigure[]{
 \begin{minipage}[t]{0.32\textwidth}
 \centering
 \includegraphics[width=\textwidth]{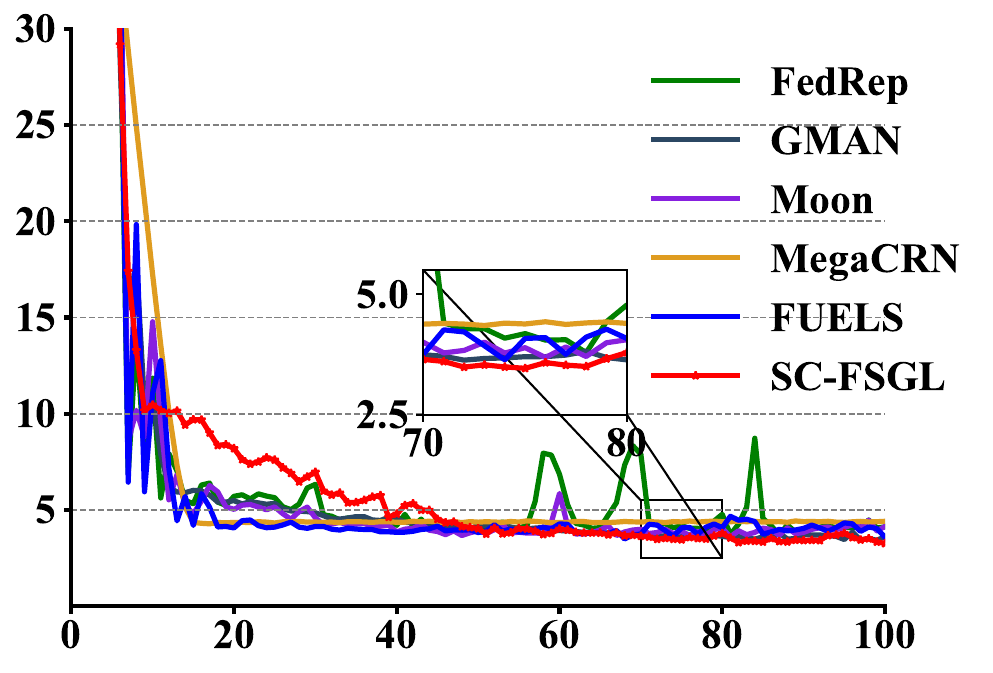}
 \end{minipage}
 }
 \subfigure[]{
 \begin{minipage}[t]{0.32\textwidth}
 \centering
 \includegraphics[width=\textwidth]{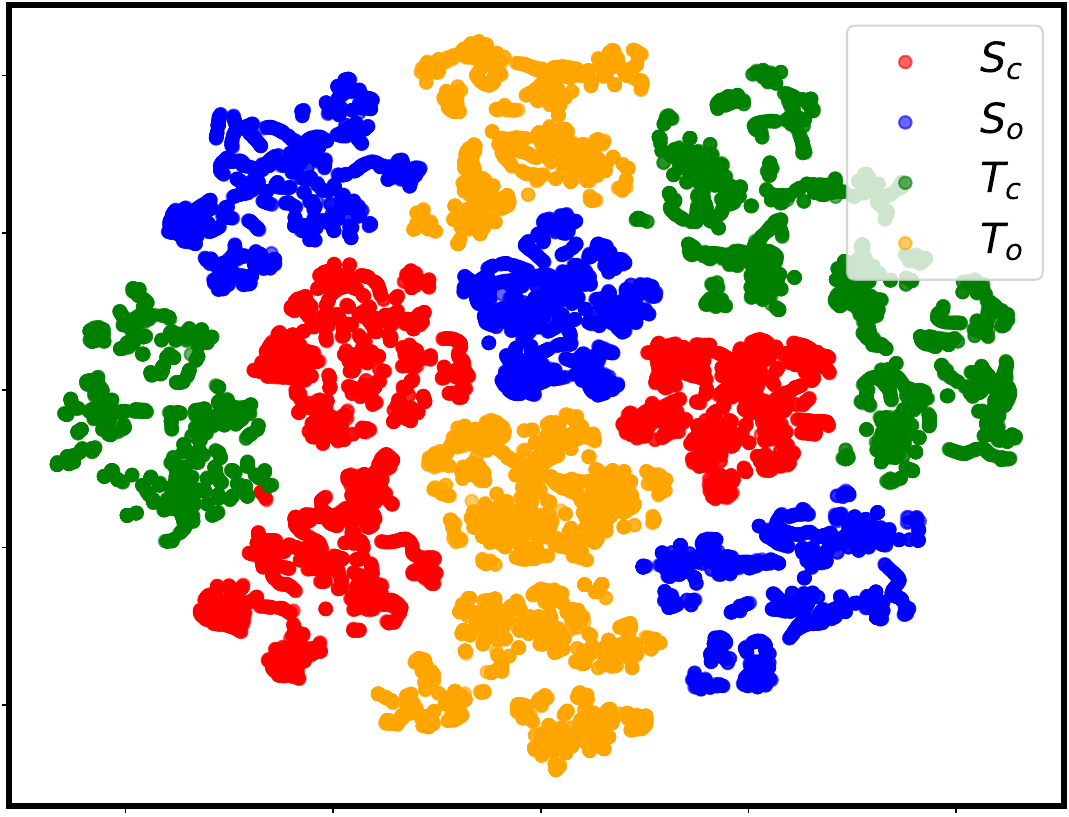}
 \end{minipage}
 }
 \caption{Performance Analysis of SC-FSGL on the PEMSD4 Dataset: (a) Prediction Results, (b) MAE Curves, and (c) t-SNE Visualization of Disentangled Causal Representations.}
 \label{fig:predict}

\end{figure*}

\begin{figure}[t]
\centering
\includegraphics[width=1\linewidth]{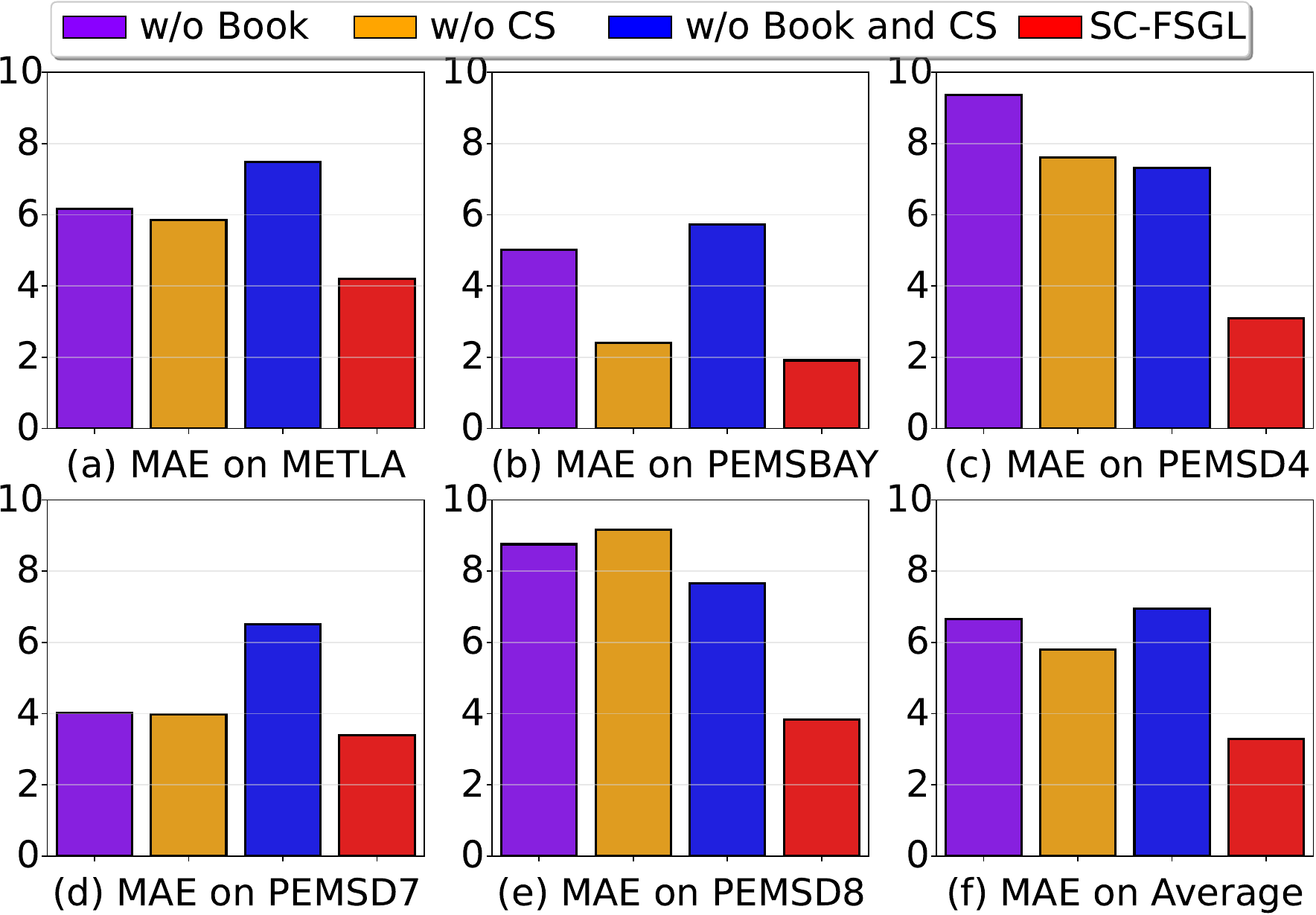}

\caption{
Ablation Study of SC-FSGL.}
\label{fig:client_performance}

\end{figure}

\subsection{Ablation Study}
In this section, we conducted ablation experiments on SC-FSGL. Using MAE as the evaluation metric, we performed ablation experiments on five datasets. We then reported the average results. In the Figure~\ref{fig:client_performance}, “w/o Book” indicates the removal of the causal codebook, “w/o CS” indicates the removal of the Conditional Separation Module, and “w/o Book and CS” indicates the simultaneous removal of both the causal codebook and Conditional Separation Modules. We observed that the SC-FSGL with the complete modules showed the lowest MAE in all variants, and this demonstrated the effectiveness of our approach. In Figure~\ref{fig:client_performance}(f), the SC-FSGL showed the lowest MAE. In contrast, the “w/o Book” showed a significant increase in MAE, indicating the importance of the causal codebook for predictive performance. The “w/o CS” variant mainly investigated whether to conduct conditional separation or not, and the results indicated that the CS contributes to the overall predictive performance of the model.

\subsection{Predictive Performance Analysis}
Figure~\ref{fig:predict}(a) shows the visual comparison of the predicted results with the ground truth on the PEMS04 dataset. Compared to other baseline methods like FedStar, FUELS, and GMAN, SC-FSGL fits the ground truth much more closely. The predicted result by SC-FSGL not only aligns well with the magnitude and trend of the actual data but also captures the temporal fluctuations and peak variations more accurately. In contrast, competing methods tend to either smooth out the signal, missing key variations, or exhibit phase shifts in temporal alignment. This demonstrates SC-FSGL’s ability to model fine-grained temporal dynamics in a federated setting. The predictive accuracy is due to the disentanglement of shared representations and client-specific representations, which can better generalize and preserve client-specific knowledge.
\subsection{MAE Trends over Communication Rounds}
Figure~\ref{fig:predict}(b) depicts the trend of MAE over the training rounds on the PEMSD4 dataset. Compared to the representative baseline methods like FedRep, Moon, GMAN, FUELS, and MegaCRN, SC-FSGL shows lower MAE throughout the training process. While some baselines suffer from a drop of MAE followed by plateaus or fluctuation, SC-FSGL shows a smooth and steady decrease, indicating the stable learning dynamics and enhanced robustness under the heterogeneous conditions. This performance can be attributed to SC-FSGL's causal disentanglement mechanism, which can separate transferable knowledge and suppress client-specific knowledge. The observed trend show cases SC-FSGL's capability to utilize global causal structures for enhanced federated prediction.
\subsection{Causal Feature Visualization via t-SNE}
To evaluate the effectiveness of SC-FSGL in disentangling causal representations, we visualize the learned features using t-SNE, as shown in Figure~\ref{fig:predict}(c). The embeddings of $\mathcal{S}_c$, $\mathcal{S}_o$, $\mathcal{T}_c$, and $\mathcal{T}_o$ are clearly separable by semantic type and show multiple compact clusters inside each category. This indicates the heterogeneity and non-stationary characteristics of the data. The shared representations ($\mathcal{S}_c$, $\mathcal{T}_c$) form tight sub-clusters, which means that the model can capture diverse and transferable patterns across clients, rather than following a single unified mode. This is further strengthened by the prototype-based alignment mechanism of the causal codebook, which promotes the representations to gather around semantic anchors. In contrast, the private representations ($\mathcal{S}_o$, $\mathcal{T}_o$) seem more scattered, allowing local heterogeneity to be more evident. The SC-FSGL can effectively discriminate transferable knowledge from client-specific factors while maintaining structural diversity, which is conducive to robust learning in heterogeneous FSTGs.

\section{Conclusion}
In this study, we propose SC-FSGL, a methodology for predicting heterogeneous FSTG data. The SC-FSGL methodology is designed to enhance the influence of shared causal relationships on model predictions, reduce the effects of client-specific causal knowledge on the global model, and decrease the granularity distance between temporal and spatial causal variables. We conduct experiments using five real-world datasets to evaluate the performance of our approach. Results demonstrate superior performance compared to prior methods on multiple baselines.

\section{Acknowledgments}
This work is supported by the National Natural Science Foundation of China under grants 62376103,  62302184, 62436003 and 62206102;  Major Science and Technology Project of Hubei Province under grant 2024BAA008.
\bibliography{aaai2026}

\clearpage

\clearpage
\appendix
\section{Appendix}
\appendix

\label{lab:appendix}

\subsection{DataSets}

 We employed five distinct clients, each utilizing different datasets characterized by varying road networks, temporal extents, and sensor counts. As depicted in Table~\ref{tab:Datasets}, \noindent\textbf{METRLA} encompasses 207 sensors from Los Angeles, spanning from Mar. 1, 2012, to Jun. 27, 2012, with a total of 34,272-time steps; \noindent\textbf{PEMSD4} encompasses 307 sensors from the San Francisco Bay Area, this dataset spans from Jan. 1, 2018, to Feb. 28, 2018, with 34,249-time steps; \noindent\textbf{PEMSD7(M)} derived from weekdays of May and June 2012 in California, this dataset includes 228 sensors and 16,969-time steps; \noindent\textbf{PEMSD8} Originate from the San Bernardino Area, this dataset contains 170 sensors and spans from Jul. 1, 2016, to Aug. 31, 2016, comprising 12,649-time steps; \noindent\textbf{PEMSBAY} consists of 325 sensors from the San Francisco Bay Area, covering the period from Jan. 1, 2017, to Jun. 30, 2017, and includes 17,833-time steps. To ensure the heterogeneity of STGs, we establish five clients, each with a different base dataset, thereby ensuring the fairness and efficacy of our experiments. We partition the data on each client into training, validation, and test sets in a ratio of 7:1:2.\\
 \noindent\textbf{Implementation Details.}
For all federated learning methods, communication rounds were set to early stop, and local training consisted of one epoch. Our implementation, developed using PyTorch, was executed on two NVIDIA 3090 GPUs for all experiments. Each dataset has unique traffic network structures and timestamps. 
 \setcounter{table}{0}
\setcounter{equation}{0}
\setlength{\tabcolsep}{3pt}
\begin{table}[htbp]
\centering
\resizebox{\columnwidth}{!}{ 
\begin{tabular}{ccccc}
\toprule
Datasets & Nodes & Data Points & Time Steps & PLACE \\ \cmidrule(r){1-1}\cmidrule(r){2-5}
METRLA & 207 & 7,094,304 & 34272 & Los Angeles \\ 
PEMSD4 & 307 & 5,216,544 & 34249 & SF Bay Area \\ 
PEMSD7(M) & 228 & 2,889,216 & 16969 & California \\ 
PEMSD8 & 170 & 3,035,520 & 12649 & SB Area \\ 
PEMSBAY & 325 & 16,937,700 & 17833 & SF Bay Area \\ 
\bottomrule
\end{tabular}
}
\caption{Here are the specific details for the five datasets, with each client possessing one dataset.}
\label{tab:Datasets}
\end{table}
 \setcounter{figure}{0}

\subsection{Baseline Methods}

To validate the effectiveness of our proposed SC-FSGL framework, we compare it against a comprehensive set of state-of-the-art baseline methods across three categories: federated learning approaches, spatio-temporal graph forecasting models, and hybrid techniques.
\begin{itemize}
  \item \textbf{FedAvg}~\cite{mcmahan2017communication}: The classic federated averaging algorithm that aggregates model weights from all clients equally. It serves as the most fundamental FL baseline, but is not designed to handle heterogeneous data or graph structures.
  
  \item \textbf{FedProx}~\cite{li2020federated}: A generalization of FedAvg that introduces a proximal term to stabilize training under data heterogeneity.
  
  \item \textbf{FedRep}~\cite{collins2021exploiting}: Separates local and global model components, allowing personalization by decoupling shared and private representations.
  
  \item \textbf{Moon}~\cite{li2021model}: A model-contrastive FL method that aligns local representations with global ones to improve generalization on non-IID data.
  
  \item \textbf{FedStar}~\cite{tan2023federated}: A structural representation alignment approach tailored for graph data, extracting structural-invariant features across clients.
  \item \textbf{GMAN}~\cite{zheng2020gman}: A graph multi-attention network that jointly models spatial and temporal dynamics through attention mechanisms.
  
  \item \textbf{MegaCRN}~\cite{jiang2023spatio}: A spatio-temporal meta-graph learning method with adaptive graph construction for traffic prediction.
  
  \item \textbf{FUELS}~\cite{liu2024personalized}: A recent federated spatio-temporal model based on contrastive learning and dual semantic alignment.
  \item \textbf{Local (no federated Algorithm):} A client-specific model trained only on local data without any communication or aggregation. This serves as the lower-bound reference for each client’s performance.
\end{itemize}
These baselines cover a wide spectrum of existing strategies, including centralized, decentralized, personalized, and contrastive approaches. By comparing against them, we demonstrate the advantages of SC-FSGL in jointly achieving accuracy, robustness, and communication efficiency under federated heterogeneous settings. 

\begin{table}[htbp]
\centering
\small
\begin{tabular}{@{} l p{6cm} @{}}
\toprule
\textbf{Baseline} & URL \\
\cmidrule(r){1-1}\cmidrule(r){2-2}
FedAvg & {https://github.com/katsura-jp/fedavg.pytorch} \\
FedProx & {https://github.com/litian96/FedProx} \\
FedRep & {https://github.com/lgcollins/FedRep} \\
Moon & 
{https://github.com/QinbinLi/MOON} \\
FedStar & 
{https://github.com/yuetan031/FedStar} \\
MegaCRN & 
{https://github.com/deepkashiwa20/MegaCRN} \\
FUELS & Self-implemented (no official code). \\
\bottomrule
\end{tabular}
\caption{Baseline methods and their GitHub links.}
\label{tab:baseline_links}
\end{table}

\subsection{Evaluation Metrics}
To quantitatively evaluate the forecasting performance of our proposed SC-FSGL framework and all baselines, we adopt three widely-used error metrics in spatio-temporal prediction tasks:

\begin{itemize}
  \item \textbf{Mean Absolute Error (MAE):} Measures the average absolute difference between predicted values and ground truth. It is defined as:
\begin{equation}
    \text{MAE} = \frac{1}{N} \sum_{i=1}^{N} | \hat{y}_i - y_i |
\end{equation}

  \item \textbf{Root Mean Squared Error (RMSE):} Penalizes large deviations more heavily by squaring the residuals before averaging. It is defined as:
\begin{equation}
    \text{RMSE} = \sqrt{ \frac{1}{N} \sum_{i=1}^{N} (\hat{y}_i - y_i)^2 }
\end{equation}

  \item \textbf{Mean Absolute Percentage Error (MAPE):} Captures relative prediction error and is scale-independent. It is defined as:
\begin{equation}
    \text{MAPE} = \frac{100\%}{N} \sum_{i=1}^{N} \left| \frac{\hat{y}_i - y_i}{y_i + \epsilon} \right|
\end{equation}
where $\epsilon$ is a small constant to prevent division by zero.
\end{itemize}

 \subsection{Hyper-parameter Studies}
We analyze the sensitivity of SC-FSGL to the hyperparameters $a$ and $b$, which control the contrastive loss $\mathcal{L}_{\text{com}}$ and the IRM regularization term $\mathcal{L}_{\text{IRM}}$, respectively. We test different combinations of $a, b \in \{0.5, 1.0, 1.5, 2.0\}$, and report MAE, RMSE, and MAPE metrics in Appendix Figure~\ref{fig.Hyper-parameter studies}. Notably, when $b=0.5$, the model exhibits a significant increase in error across all metrics, indicating that removing or weakening the IRM regularization leads to degraded generalization under client heterogeneity. In contrast, setting $a = b = 1$ consistently yields optimal performance. This confirms that both the contrastive alignment and the IRM-inspired invariance objective contribute critically to the stability and predictive effectiveness of SC-FSGL. These results empirically validate our design choice of incorporating the IRM loss to ensure representation robustness across heterogeneous client environments.

We analyzed the model hyperparameters, denoted as $a$ and $b$. The $a$ represents the weight of the comparative loss $L_{com}$ in the loss function, while $b$ represents the weight of the predictive loss $L_{pred}$. They respectively control the influence of the comparative loss $L_{com}$ and the predictive loss $L_{pred}$ on the model’s convergence. We experimented with hyperparameters $a=(0.5,1.0,1.5,2.0)$ and $b=(0.5,1.0,1.5,2.0)$, and the experimental results are illustrated in Figure~\ref{fig.Hyper-parameter studies}. When $a=b=1$, the MAE, MAPE and RMSE are optimal. This is because both the comparative loss $L_{com}$ and the predictive loss $L_{pred}$ contribute equally to the predictive results. A small value of $b$ would make the model more inclined to bring shared spatial causal knowledge and temporal causal knowledge closer together, neglecting the connection between the predicted results and the ground truth. Therefore, to ensure fairness in the contributions of $L_{com}$ and $L_{pred}$ to the model, we conducted experiments with $a=b=1$.
\begin{figure}[h]

 \centering
 \subfigure[MAE]{
 \centering
 \begin{minipage}[t]{0.31\columnwidth}
 \includegraphics[width=1\columnwidth]{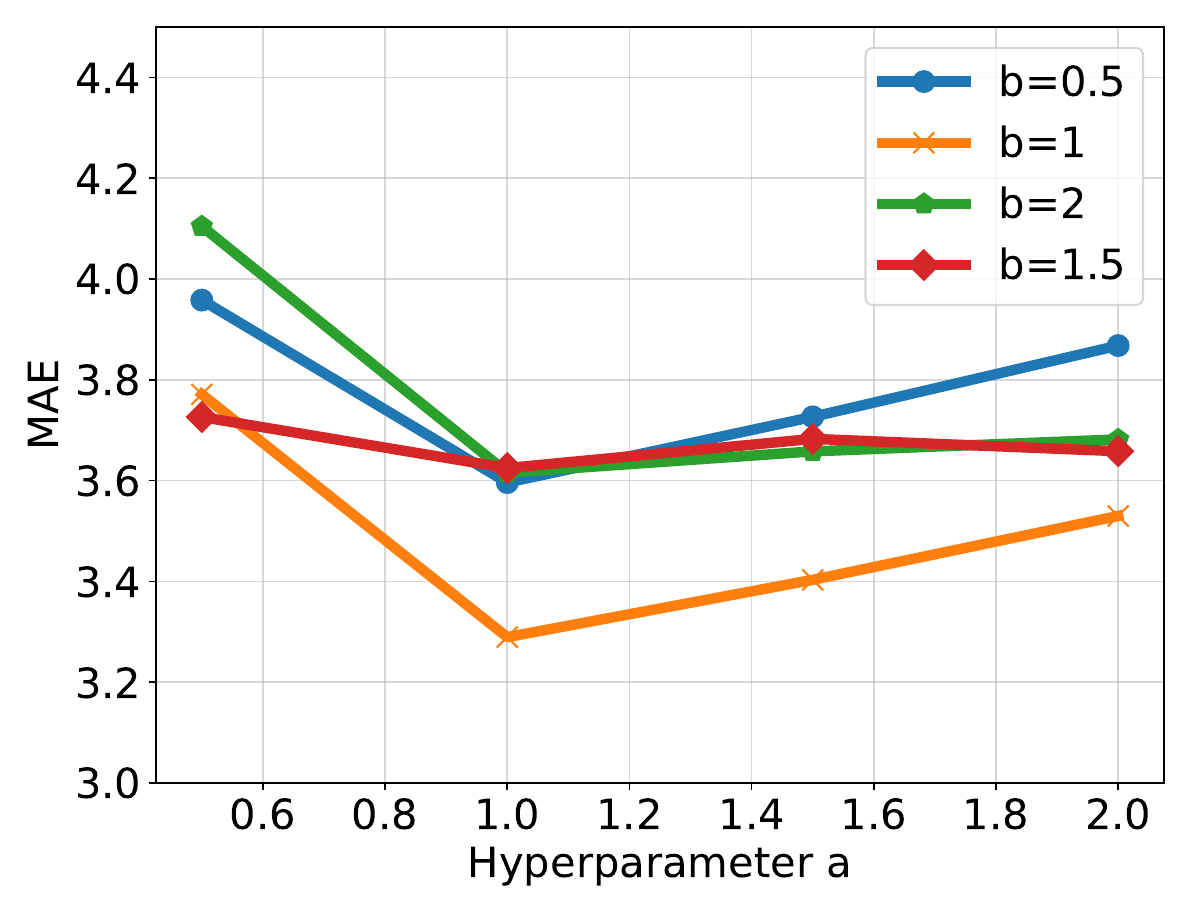}
 \end{minipage}%
 }
 \subfigure[RMSE]{
 \centering
 \begin{minipage}[t]{0.31\columnwidth}
 \includegraphics[width=1\columnwidth]{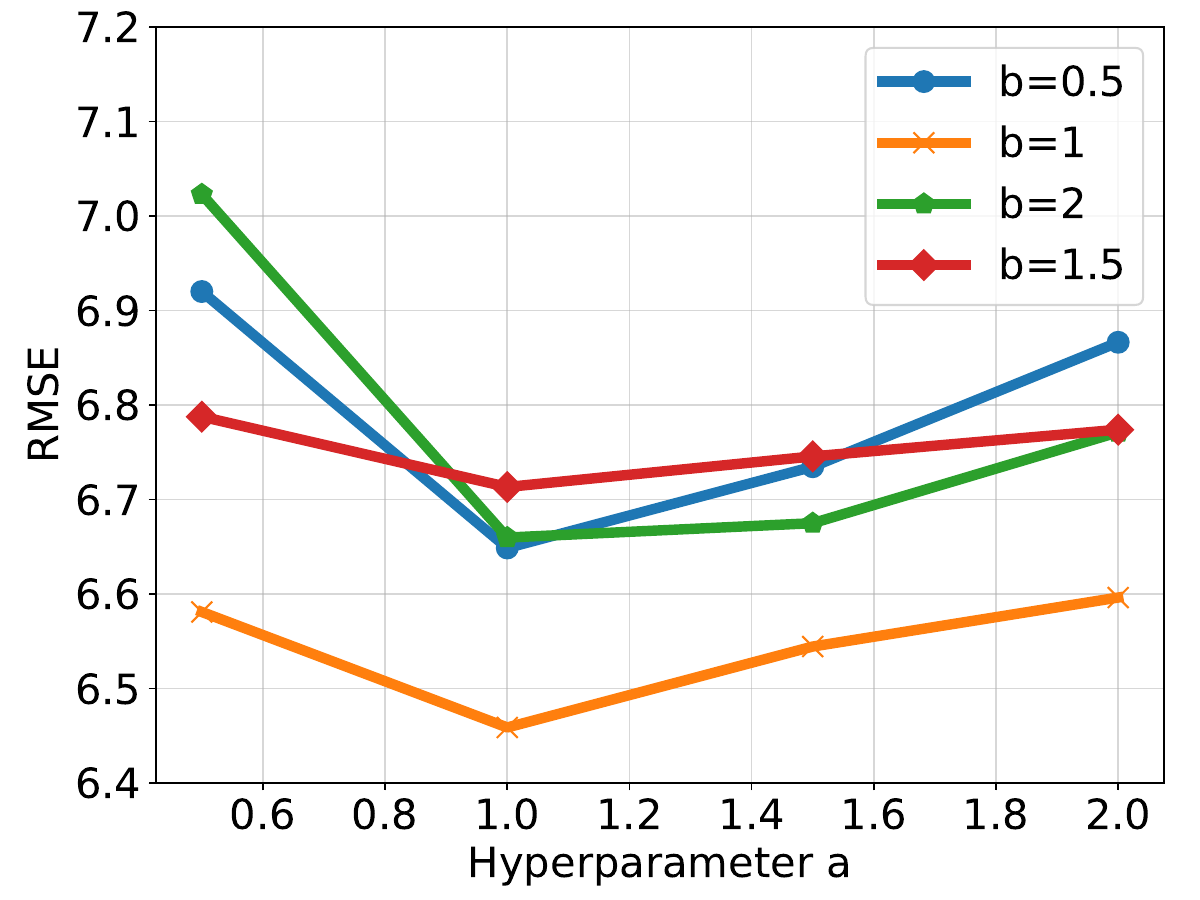}
 \end{minipage}%
 }
 \subfigure[MAPE]{
 \centering
 \begin{minipage}[t]{0.31\columnwidth}
 \includegraphics[width=1\columnwidth]{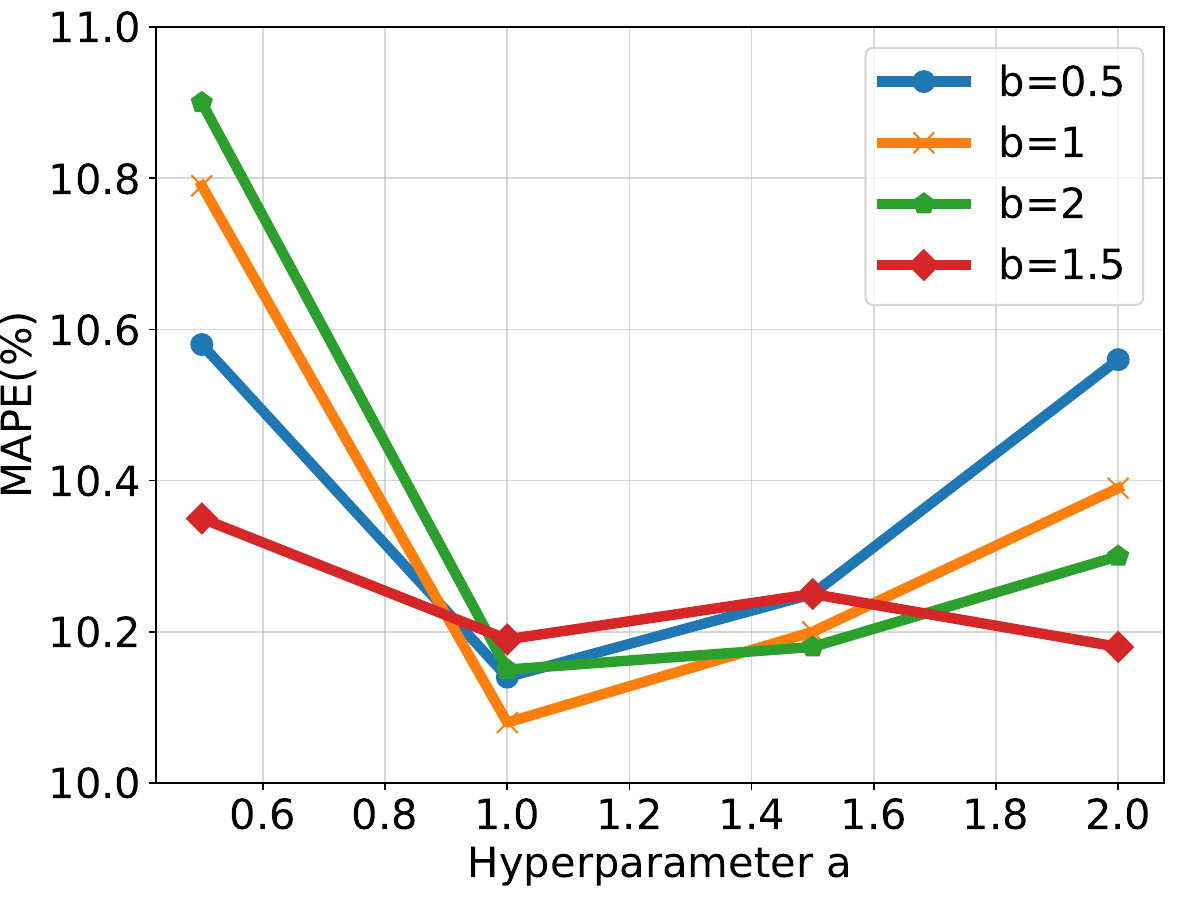}
 \end{minipage}%
 }
 \centering

 \caption{Hyper-parameter studies about $a$ and $b$.}
 \label{fig.Hyper-parameter studies}

\end{figure}

\subsection{Evolution of Latent Space via Communication Rounds}
\begin{figure}[t]

 \centering
 \subfigure[Round=1]{
 \centering
 \begin{minipage}[t]{0.31\columnwidth}
 \includegraphics[width=1\columnwidth]{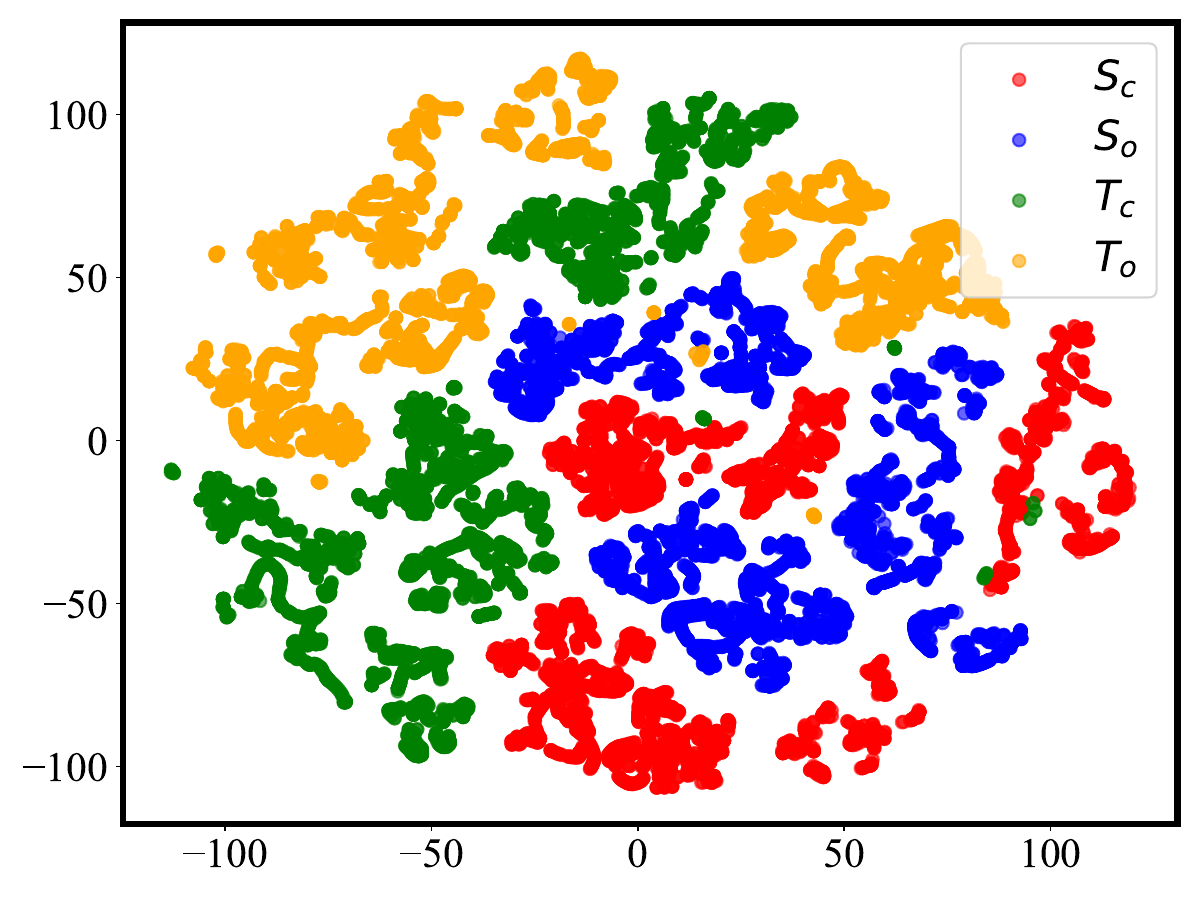}
 \end{minipage}%
 }
 \subfigure[Round=5]{
 \centering
 \begin{minipage}[t]{0.31\columnwidth}
 \includegraphics[width=1\columnwidth]{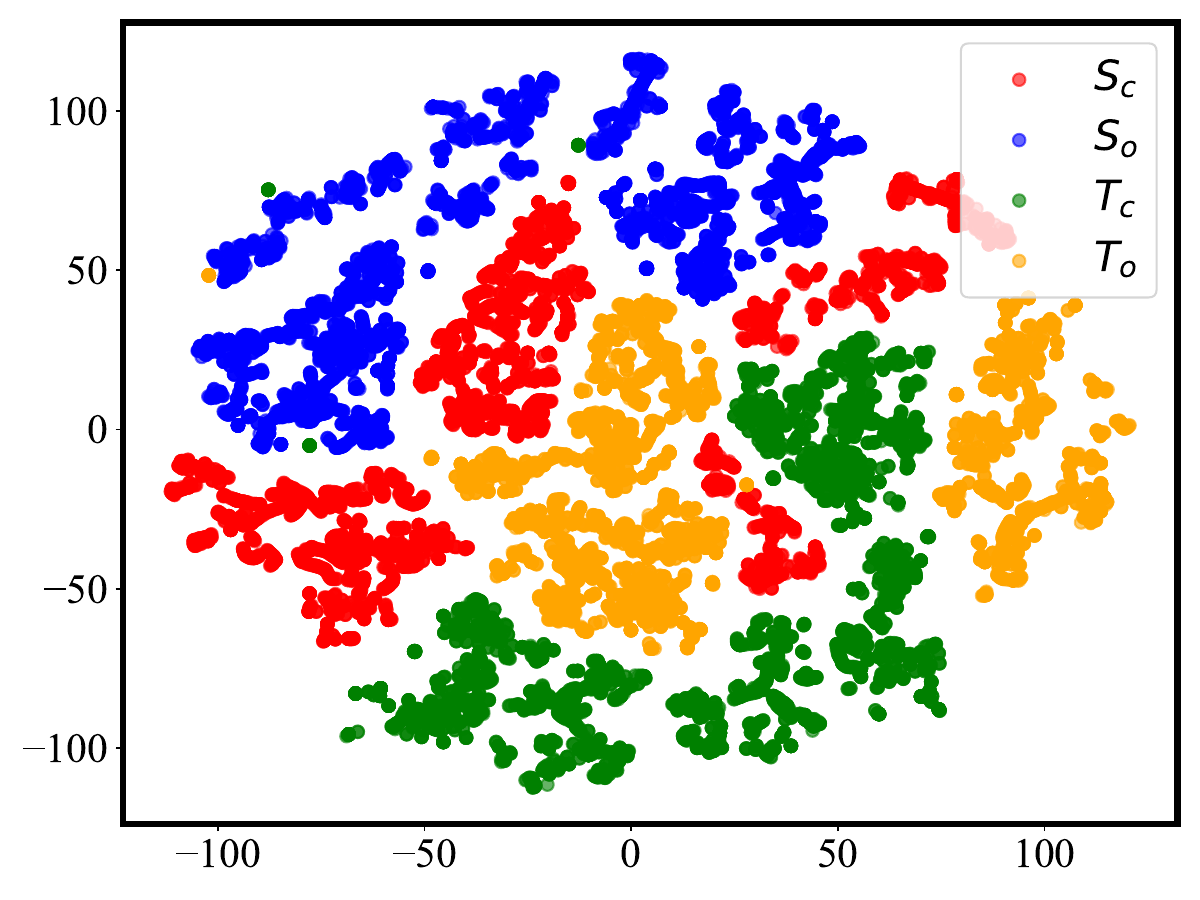}
 \end{minipage}%
 }
 \subfigure[Round=20]{
 \centering
 \begin{minipage}[t]{0.31\columnwidth}
 \includegraphics[width=1\columnwidth]{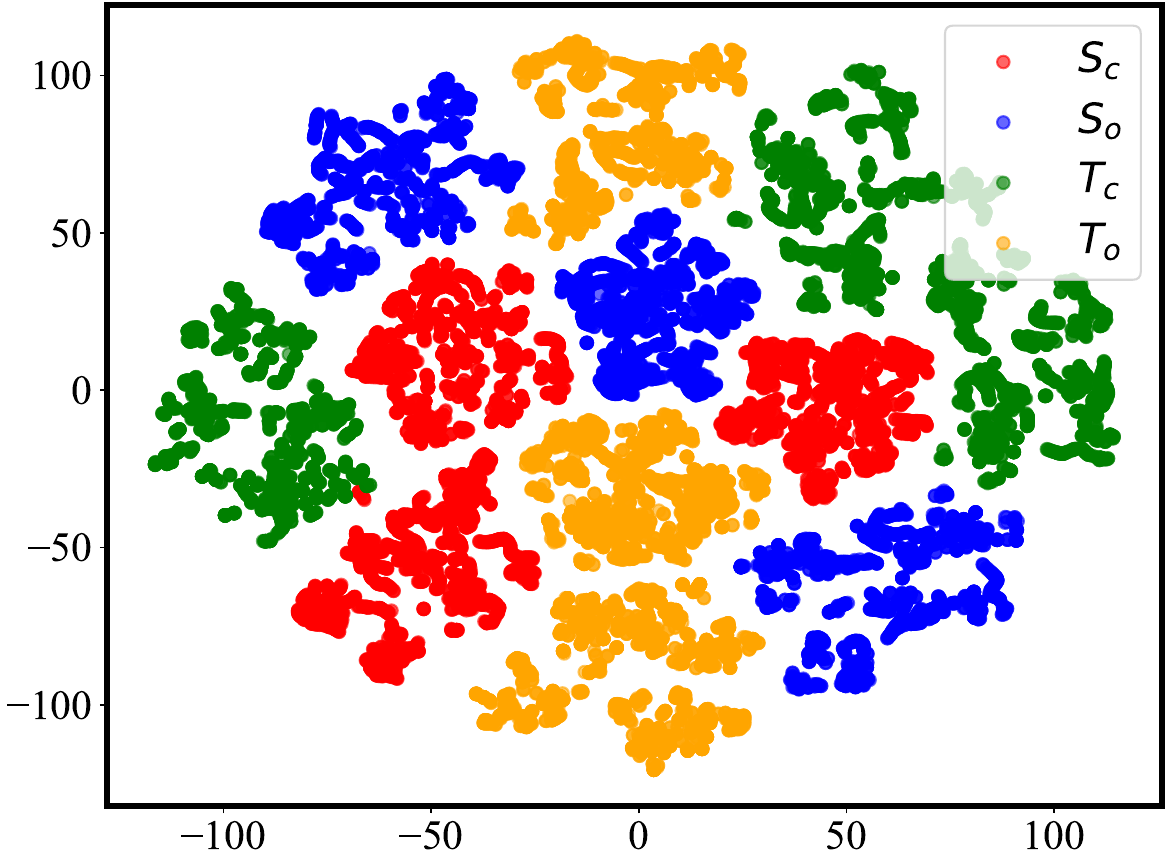}
 \end{minipage}%
 }
 \centering

 \caption{t-SNE Visualization of Representations Across Communication Rounds.}
 \label{fig.tsne over rounds}

\end{figure}

To investigate how the learned causal representations evolve over the course of federated training, we conduct a t-SNE visualization of latent features at different communication rounds. As shown in Figure~\ref{fig.tsne over rounds}, we visualize the shared and client-specific spatio-temporal representations ($\mathbf{S}_c$, $\mathbf{S}_o$, $\mathbf{T}_c$, $\mathbf{T}_o$) at Round 1, Round 5, and Round 20. At Round 1 (Figure~\ref{fig.tsne over rounds}a), initial clusters corresponding to $\mathbf{S}_c$, $\mathbf{S}_o$, $\mathbf{T}_c$, and $\mathbf{T}_o$ have already begun to emerge, indicating that the model’s disentanglement mechanism starts to take effect early. However, these clusters remain relatively scattered and partially overlapping, suggesting that shared and client-specific representations are not yet clearly separated. By Round 5 (Figure~\ref{fig.tsne over rounds}b), the separation becomes more apparent shared representations ($\mathbf{S}_c$, $\mathbf{T}_c$) show tighter grouping, while client-specific ones ($\mathbf{S}_o$, $\mathbf{T}_o$) exhibit increased dispersion. At Round 20 (Figure~\ref{fig.tsne over rounds}c), the latent space demonstrates well-formed and compact clusters with clear semantic boundaries. The shared features are consistently centralized, reflecting successful alignment across clients, whereas the private features remain more distributed, capturing local heterogeneity. This progressive refinement highlights SC-FSGL’s ability to evolve semantically meaningful representations over communication rounds. It confirms that the Conditional Separation Module and contrastive codebook effectively guide the model toward disentangling transferable causal knowledge from client-specific patterns, ultimately enhancing generalization in federated spatio-temporal settings.

\subsection{RMSE and MAPE Curves of SC-FSGL}
\begin{figure}[h]
 \centering
 \subfigure[RMSE]{
 \begin{minipage}[t]{0.47\columnwidth}
 \centering
 \includegraphics[width=\textwidth]{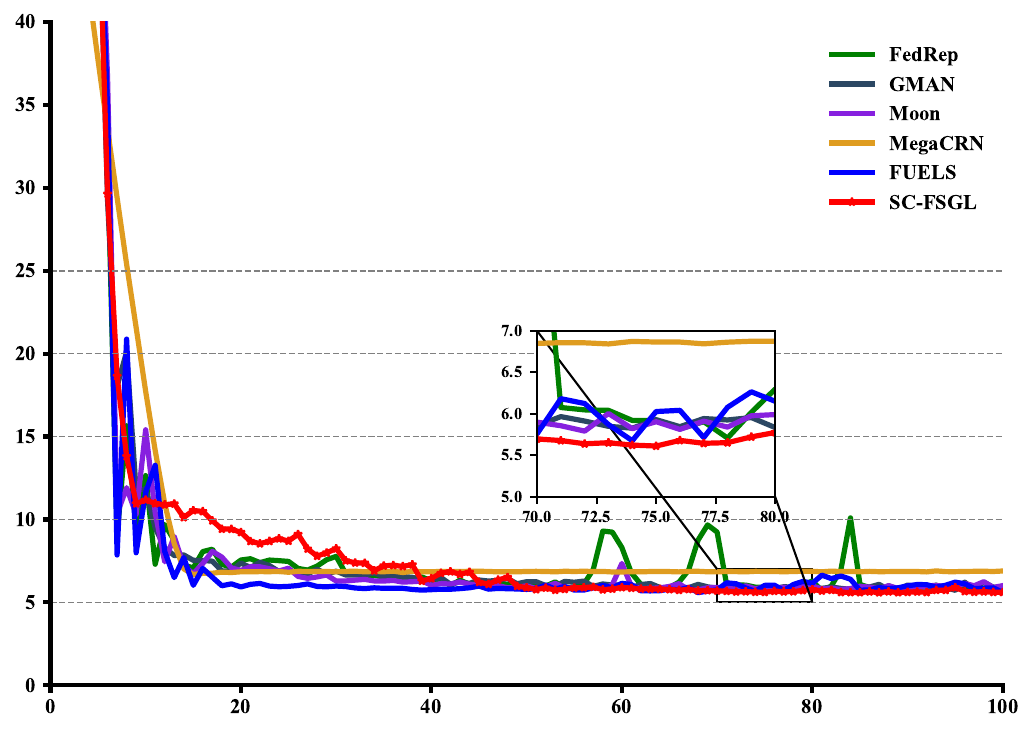}
 \end{minipage}
 }
 \hfill
 \subfigure[MAPE]{
 \begin{minipage}[t]{0.47\columnwidth}
 \centering
 \includegraphics[width=\textwidth]{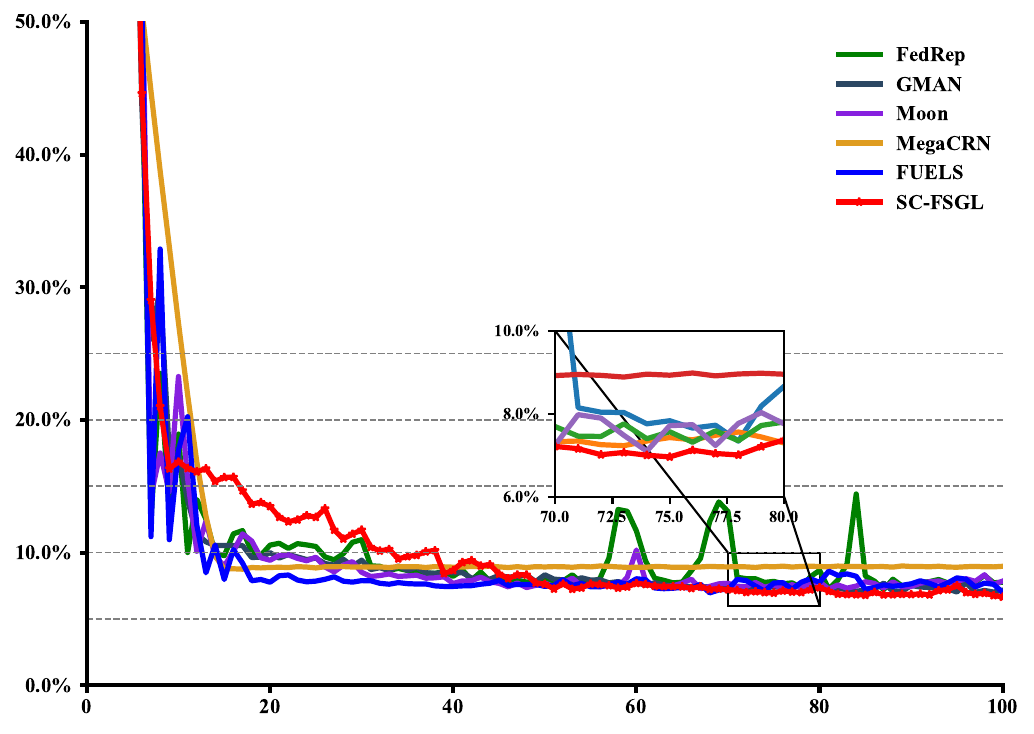}
 \end{minipage}
 }

 \caption{The RMSE and MAPE Curves of SC-FFSGL.}
 \label{fig:RMSE&MAPE}
\end{figure}

Figure~\ref{fig:RMSE&MAPE}(a) and~\ref{fig:RMSE&MAPE}(b) illustrate the evolution of RMSE and MAPE over communication rounds on the PEMSD4 dataset. SC-FSGL consistently maintains the lowest RMSE and MAPE compared with all baseline methods throughout the training process. Notably, methods such as FedRep, Moon, and MegaCRN exhibit considerable fluctuations or plateaus, indicating unstable learning under heterogeneous conditions. In contrast, SC-FSGL shows a steady and smooth decline in both RMSE and MAPE, reflecting its robust generalization capability and effective suppression of client-specific noise. The superior performance is attributed to SC-FSGL’s ability to disentangle transferable causal features and align them via the causal codebook, which not only improves prediction accuracy but also ensures stability across rounds. These results highlight the model’s effectiveness in achieving accurate and reliable spatio-temporal forecasting under heterogeneous federated settings.

\subsection{Prediction Visualization Across Clients}
To further demonstrate the effectiveness of SC-FSGL, we visualize the prediction results on two representative clients: PEMSD4 and PEMSD8, as illustrated in Figure~\ref{fig:predict}(a) and Figure~\ref{fig:predict}(b), respectively. The curves represent predicted traffic speed over time, compared against ground truth values. In both datasets, SC-FSGL yields predictions that closely follow the ground truth trajectory, particularly in capturing both the sharp rises and falls during peak and off-peak hours. This shows SC-FSGL’s strength in modeling complex temporal dynamics and adapting to local variations through its disentangled causal learning mechanism. In contrast, FedStar exhibits significant prediction bias and smoothing, consistently failing to capture the actual peak values, especially in periods of rapid change. GMAN, although designed for centralized STG learning, struggles under federated heterogeneous settings and demonstrates noticeable prediction lag. FUELS, a recent contrastive-based federated method, performs better than most baselines but still underestimates the variability in traffic speed, especially in PEMSD8. Finally, the results validate that SC-FSGL’s causal separation and codebook alignment mechanisms enable it to achieve more stable and accurate forecasts across diverse and temporally heterogeneous clients. Its predictions exhibit reduced drift, better alignment with ground truth, and superior generalization in heterogeneous FSTGs scenarios. 
\begin{figure}[h]
 \centering
 \subfigure[PEMSD4]{
 \begin{minipage}[t]{0.47\columnwidth}
 \centering
 \includegraphics[width=\textwidth]{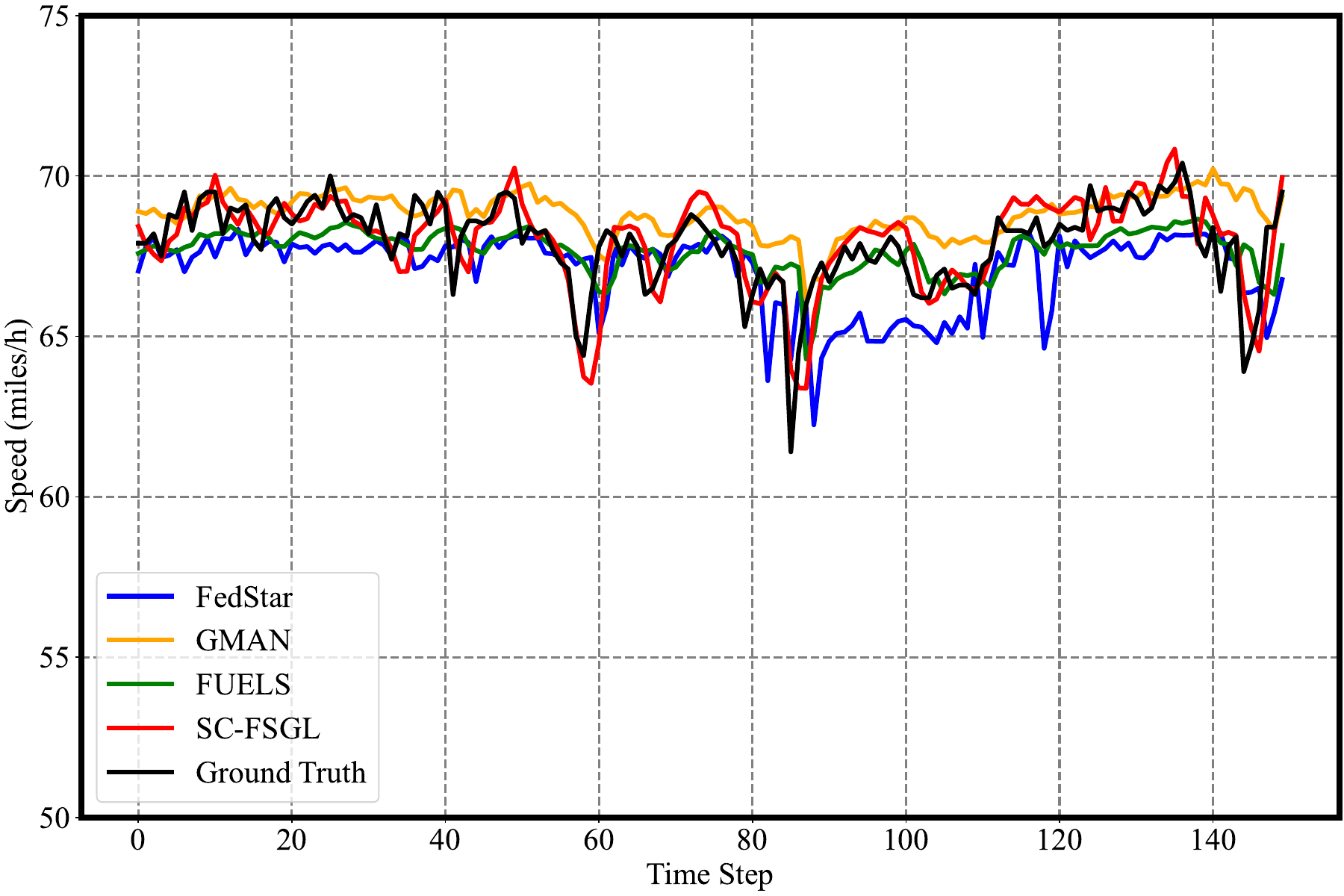}
 \end{minipage}
 }
 \hfill
 \subfigure[PEMSD8]{
 \begin{minipage}[t]{0.47\columnwidth}
 \centering
 \includegraphics[width=\textwidth]{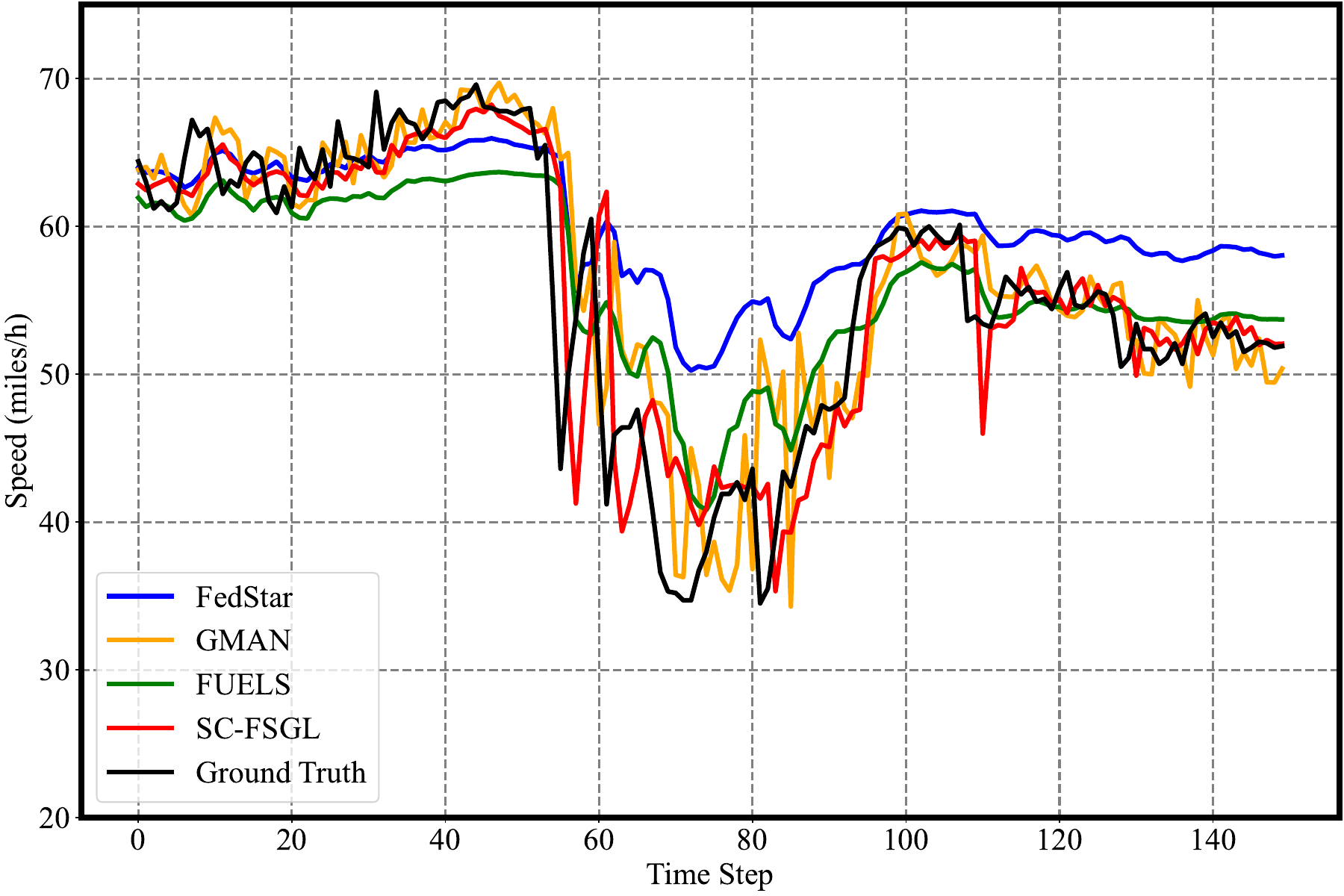}
 \end{minipage}
 }

 \caption{Visualization of Prediction Performance on Heterogeneous Clients.}
 \label{fig:predict_client}
\end{figure}

\begin{algorithm}[htbp]
\caption{SC-FSGL Framework}
\label{alg:client}
\begin{algorithmic}[1]
\STATE \textbf{Input:} $G^k_{t}=\{X^k_{t},A_t^k\}$, learning rate $\eta$, predict time steps $\beta$, Spatial Vector $SE_k$, Round $R$
\STATE \textbf{Output:} $\mathcal{Y}_t^k=\{(X^k_{t+1},X^k_{t+2},...,X^k_{t+\beta}), A_t^k\}$

\STATE \textbf{Server of SC-FSGL:}
\STATE Randomly initialize server causal codebook $\delta_{G,r=0}$
\FOR{each communication round $r = 1$ to $R$}
  \FOR{each client $k$ in parallel}
    \STATE $\delta_{k,r+1} \gets \text{ClientUpdate}(\delta_{G,r})$
  \ENDFOR
  \STATE $\delta_{G,r+1} \gets \frac{1}{K}\sum\limits_{i=1}^K \delta_{k,r+1}$
\ENDFOR

\STATE \textbf{ClientUpdate of SC-FSGL:}
\FOR{each client $k$ in parallel}
  \STATE Download $\delta_{G,r}$ from server
  \STATE $\mathcal{T}^k_t\gets MLP(\{t-\gamma,t-\gamma+1,...,t\})$
  \STATE $H_{t} \gets MLP(\mathcal{G}_t)$
  \STATE $STE_{his} \gets \textbf{concat}[\mathcal{T}^k_{t_{his}}, SE]$
  \STATE $STE_{pred} \gets \textbf{concat}[\mathcal{T}^k_{t_{pred}}, SE]$

  \FOR{$l = 1$ to $L$}
    \STATE Calculate $\mathcal{S}^k_t$ with $\Phi_s$($H_t$), and $\mathcal{T}^k_t$ with $\Phi_t$($STE_{his}$)
    \STATE Obtain $\mathcal{S}^k_{t,c}, \mathcal{S}^k_{t,o}, \mathcal{T}^k_{t,c}, \mathcal{T}^k_{t,o}$ via SCS, TCS
    \STATE $W_{Q_t}\gets Query(\mathcal{T}^k_{t,c})$, $W_{Q_s}\gets Query(\mathcal{S}^k_{t,c})$
    \STATE $\alpha^s_{j}\gets W_{Q_s} * \delta_{j}$, $\alpha^t_{j} \gets W_{Q_t} * \delta_{j}$
    \STATE $\mathcal{S}^k_{pos}, \mathcal{S}^k_{neg} \gets \alpha^s_{j}, \delta_{j}$
    \STATE $\mathcal{T}^k_{pos}, \mathcal{T}^k_{neg} \gets \alpha^t_{j}, \delta_{j}$
    \STATE $L_{com} \gets \mathcal{S}^k_{pos}, \mathcal{S}^k_{neg}, \mathcal{T}^k_{pos}, \mathcal{T}^k_{neg}$
    \STATE $\mathcal{R}^k_t \gets \textbf{concat}[\mathcal{T}^k_{t,o}, \mathcal{T}^k_{t,c}]$, $\textbf{concat}[\mathcal{S}^k_{t,o}, \mathcal{S}^k_{t,c}]$
  \ENDFOR

  \FOR{$l = 1$ to $L$}
    \STATE Calculate $\mathcal{S}^k_t$ with $\Phi_s$($\mathcal{R}_t$), $\mathcal{T}^k_t$ with $\Phi_t$($STE_{pred}$)
    \STATE Obtain decomposed components via SCS, TCS
    \STATE $W_{Q_t} \gets Query(\mathcal{T}^k_{t,c})$, $W_{Q_s} \gets Query(\mathcal{S}^k_{t,c})$
    \STATE $\alpha^s_{j}\gets W_{Q_s}*\delta_{k,r}$, $\alpha^t_{j} \gets W_{Q_t}*\delta_{k,r}$
    \STATE $\mathcal{S}^k_{pos}, \mathcal{S}^k_{neg} \gets \alpha^s_{j}, \delta_{j}$
    \STATE $\mathcal{T}^k_{pos}, \mathcal{T}^k_{neg} \gets \alpha^t_{j}, \delta_{j}$
    \STATE $L_{com} \gets \mathcal{S}^k_{pos}, \mathcal{S}^k_{neg}, \mathcal{T}^k_{pos}, \mathcal{T}^k_{neg}$
    \STATE $\mathcal{Y}^k_t \gets \textbf{concat}[\mathcal{T}^k_{t,o}, \mathcal{T}^k_{t,c}], \textbf{concat}[\mathcal{S}^k_{t,o}, \mathcal{S}^k_{t,c}]$
  \ENDFOR

  \STATE Calculate $L_{pred} \gets \mathcal{Y}^k_{t,true}, \mathcal{Y}^k_t$
  \STATE $\delta_{k,r+1} \gets$ encoder's $\delta_{j}$
  \STATE Calculate $L_{local} \gets L_{com} + L_{pred}+L_{IRM}$
  \STATE Send encoder causal codebook $\delta_{k,r+1}$ to Server
\ENDFOR
\end{algorithmic}
\end{algorithm}

\subsection{Algorithm}
In this section, we will present the algorithm of SC-FSGL. SC-FSGL’s local model uses an encoder-decoder architecture to predict future moments based on historical spatio-temporal data, while the global model aggregates domain-invariant knowledge from local causal codebook transmissions. The complete SC-FSGL algorithm can be divided into the following two main parts:

\textbf{The ClientUpdate algorithm section is illustrated in Algorithm~\ref{alg:client}}.

\textbf{Spatio-temporal Embedding}: For spatial dimension information, each client utilizes a fully connected layer to transform historical data from $ G^k_t=(X^k_t,A^k_t)$ to hidden state $H_t \in \mathbb R^{\vert V\vert \times D}$. Where $D$ represents the output dimension of the MLP and $k$ represents $k$-th client. For temporal dimension information, each client concatenates spatial embedding SE and the time series $\mathcal{T}^k_{t_{his}}=\{t-\gamma, t-\gamma+1, ..., t\}$ with $\mathcal{T}^k_{t_{pred}}=\{t+1, t+2, ...,t+\beta\}$ through a MLP, ultimately obtaining STEmbedding $STE_{his}=\textbf{concat}[\mathcal{T}^k_{t_{his}}, SE]$ and $STE_{pred}=\textbf{concat}[\mathcal{T}^k_{t_{pred}}, SE]$. The specific steps are detailed in Algorithm~\ref{alg:client} lines 3 to 5.

\textbf{Encoder and Decoder}: We input the historical traffic conditions into the encoder-decoder structure to predict future traffic conditions. The specific steps are as follows: Inputting $H_t$ into the spatial attention network $\Phi_s$ of the encoder to obtain spatial representation. Inputting spatio-temporal representation $STE_{his}$ into the time attention network $\Phi_t$ of the encoder to obtain time representation $\mathcal{T}^k_t$. These are then separately passed through spatial conditional segregation modules~(SCS) to obtain $\mathcal{S}^k_{t,c}$, $\mathcal{S}^k_{t,o}$ and through Temporal conditional segregation modules~(TCS) to obtain $\mathcal{T}^k_{t,o}$,$\mathcal{T}^k_{t,c}$. By calculating the similarity between causal codebook $\delta$ and $\mathcal{S}^k_{t}$ ,$\mathcal{T}^k_{t}$, we select the most similar item as $\mathcal{S}^k_{pos}$ and $\mathcal{T}^k_{pos}$, and the second most similar as $\mathcal{S}^k_{neg}$ and $\mathcal{T}^k_{neg}$. Subsequently, we use a loss function to minimize the distance between $\mathcal{S}^k_{pos}$ amd $\mathcal{T}^k_{pos}$ and maximize the distance between $\mathcal{S}^k_{pos}$, $\mathcal{T}^k_{pos}$ and $\mathcal{S}^k_{neg}$, $\mathcal{T}^k_{neg}$. Finally, we input $\textbf{concat}[\mathcal{T}^k_{t,o},\mathcal{T}^k_{t,c}]$ and $\textbf{concat}[\mathcal{S}^k_{t,o},\mathcal{S}^k_{t,c}]$ into a prediction head to obtain the spatio-temporal representation $\mathcal{R}^k_t$. The decoder structure is similar to the encoder structure, with the difference being the input of $STE_{pred}$ and $\mathcal{R}^k_t$. For specifics, refer to lines 8 to 16 of Algorithm~\ref{alg:client} for the encoder and lines 20 to 28 of Algorithm~\ref{alg:client} for the decoder.\\

 \textbf{The server-side algorithm specifics can be found in Algorithm~\ref{alg:client}}.

Aggregation algorithm: After each epoch of local training, we aggregate causal codebook $\delta_k(k\in K)$ to transmit domain-invariant knowledge from different clients. Subsequently, we send the aggregated global causal codebook $\delta_G$ to each client as the model parameters for the next round of causal codebook.

\subsection{Privacy Discussion} Compared to traditional FedAvg and its derivative algorithms, SC-FSGL exchanges causal codebooks between the server and clients rather than model parameters. First, the causal codebook inherently protects data privacy because it learns an intermediate distribution that is closest to the distributions of $\mathcal{S}^k_{t,c}$ and $\mathcal{T}^k_{t,c}$, a process that is irreversible. Furthermore, no raw data or local models are transmitted during this process, meaning attackers cannot target the model or reconstruct the original data from the causal codebook. Finally, the causal codebook can be combined with various privacy-preserving techniques to further enhance system privacy. Unlike prototypes, the causal codebook does not disclose any class or temporal information, preventing potential data leakage.
\subsection{Limitations and Future Works}
While SC-FSGL has shown strong performance across various heterogeneous spatio-temporal scenarios, some secondary aspects remain to be improved. For instance, the causal codebook is designed with a fixed size and global scope, which may limit its expressiveness when facing highly diverse or evolving client semantics. Additionally, although the framework employs soft interventions to approximate causal effects, it does not explicitly model temporal causal drift, which may affect long-term forecasting under dynamic environments. Another limitation lies in the experimental scope, as evaluations are conducted mainly on traffic datasets; generalizing SC-FSGL to other domains such as environmental sensing or smart grids is a valuable future direction. Addressing these factors by incorporating adaptive prototype structures, dynamic causal modeling, and cross-domain validation would further enhance the robustness and applicability of the proposed framework.
\subsection{Communication Efficiency of SC-FSGL}

In FL, communication efficiency is a critical concern due to the high cost of transmitting large-scale model parameters across distributed clients. Unlike most prior works that transmit the full local model (e.g., encoder, decoder, or GNN layers) during each communication round, our proposed SC-FSGL framework significantly reduces communication overhead by requiring only the exchange of the causal codebook.

Specifically, each client uploads a lightweight causal codebook $\delta^k \in \mathbb{R}^{\phi \times d}$ after local training, where $\phi$ is the number of causal prototypes and $d$ is the feature dimension. The server aggregates these prototypes via averaging to form a new global codebook $\delta^{\text{global}}$, which is then broadcasted back to clients. No other model parameters (e.g., $\theta$, $\omega$) are transmitted, as local models remain private and personalized.

This design brings two major benefits:
\begin{itemize}
  \item \textbf{Reduced Bandwidth Cost:} The total communication per round scales linearly with the size of the codebook ($\phi \times d$), which is orders of magnitude smaller than transmitting full neural network weights.
  \item \textbf{Privacy by Design:} Since only abstract, non-instance-level causal prototypes are communicated, SC-FSGL inherently preserves model and data privacy, aligning with the principles of secure federated learning.
\end{itemize}
In our experiments, we set $\phi=32$ and $d=64$, resulting in less than 8KB of communication per round per client a negligible cost compared to typical GNN model updates. This demonstrates that SC-FSGL is both scalable and communication-efficient for large-scale federated spatio-temporal learning tasks.

\subsection{Proof of Interventional Distributions}
For Equation~\ref{eq.adj_T}, we integrate the two estimated values, $ P(\mathcal{Y}_t^k | \textbf{do}(\mathcal{T}_{t,c}^k)) $ and $ P(\mathcal{Y}_t^k | \textbf{do}(\mathcal{S}_t^k)) $. The complete derivation process is as follows:

\begin{align}
  \label{eq:18}
\begin{split}
  P(\mathcal{Y}_t^k | \textbf{do}(\mathcal{T}_{t,c}^k)) &= \sum P(\mathcal{Y}_t^k | \textbf{do}(\mathcal{T}_{t,c}^k), \mathcal{T}_{t,o}^k) P(\mathcal{T}_{t,o}^k | \textbf{do}(\mathcal{T}_{t,c}^k))\\
  &= \sum P(\mathcal{Y}_t^k | \textbf{do}(\mathcal{T}_t^k)) P(\mathcal{T}_{t,c}^k)\\
  &= \sum P(\mathcal{T}_{t,o}^k) \sum P(\mathcal{Y}_t^k | \textbf{do}(\mathcal{T}_t^k), \mathcal{S}_t^k) \\&~~~~~P(\mathcal{S}_t^k | \textbf{do}(\mathcal{T}_t^k))\\
  &= P(\mathcal{S}_t^k) \sum P(\mathcal{Y}_t^k | \mathcal{G}_t^k) P(\mathcal{S}_t^k)
\end{split}
\end{align}
Similarly, for $P(\mathcal{Y}_t^k | \textbf{do}(\mathcal{S}_{t,c}^k)$, we have:
\begin{align}
  \label{eq:19}
\begin{split}
P(\mathcal{Y}_t^k | \textbf{do}(\mathcal{S}_{t,c}^k)) &= \sum P(\mathcal{Y}_t^k | \textbf{do}(\mathcal{S}_{t,c}^k), \mathcal{S}_{t,o}^k) P(\mathcal{S}_{t,o}^k | \textbf{do}(\mathcal{S}_{t,c}^k))\\ 
&= \sum P(\mathcal{Y}_t^k | \textbf{do}(\mathcal{S}_t^k)) P(\mathcal{S}_{t,c}^k)\\
&= \sum P(\mathcal{S}_{t,o}^k) \sum P(\mathcal{Y}_t^k | \textbf{do}(\mathcal{S}_t^k), \mathcal{T}_t^k) \\
&~~~~~~P(\mathcal{T}_t^k | \textbf{do}(\mathcal{S}_t^k))\\
&= P(\mathcal{T}_t^k) \sum P(\mathcal{Y}_t^k | \mathcal{G}_t^k) P(\mathcal{T}_t^k)
\end{split}
\end{align}

\subsection{Time Complexity Analysis.}
In this section, we evaluate the time complexity of the SC-FSGL model, which consists of four main components: the feature extractor, the Conditional Separation Module, the Causal codebook, and the prediction head. The feature extractor in SC-FSGL uses an attention mechanism, with a time complexity of $O\left(\sum_{l=1}^{L_{feat}} \gamma^2 \cdot D^l_{in} \cdot D^l_{out}\right)$, where $\gamma$ represents the time step, $D^l_{in}$ is the input feature dimension of the $l$-th layer, $D^l_{out}$ is the output feature dimension, and $L_{feat}$ is the number of layers in the feature extractor. Additionally, since SC-FSGL employs node2vec as the structure extractor, its precomputation introduces an extra complexity of $O(|V|d + |E|)$ per iteration for graph traversal, plus $O(|V| r l)$ for $r$ random walks of length $l$, and $O(|V|d)$ for embedding training. The Conditional Separation Module includes two learnable mask modules, $\mathcal{M}_t^{k,c}$ and $\mathcal{M}_t^{k,o}$, each with a time complexity of $O(\gamma \cdot D_{in} \cdot D_{out})$, where $D_{in}$ and $D_{out}$ denote the input and output feature dimensions, respectively. For the Causal codebook, the initial time complexity is $O(P \cdot F)$, where $P$ is the number of Causal book items, and $F$ is the feature dimension of each item. The time complexity of the prediction head is similar to that of a multi-layer perceptron (MLP), specifically $O\left(\sum_{l=1}^{L_{pred}} D^l_{in} \cdot D^l_{out}\right)$,
where $L_{pred}$ is the total number of layers in the prediction head, and $D^l_{in}$ and $D^l_{out}$ are the input and output feature dimensions of the $l$-th layer. Since the feature extractor and Conditional Separation Module appear together, their time complexities are multiplied by 2. Thus, the overall time complexity of the model is $O\left(\sum_{l=1}^{L_{pred}} D^l_{in} \cdot D^l_{out}\right) + O(P \cdot F) + 2 \cdot O(\gamma \cdot D_{in} \cdot D_{out}) + 2 \cdot O\left(\sum_{l=1}^{L_{feat}} \gamma^2 \cdot D^l_{in} \cdot D^l_{out}\right)+O(\vert V\vert d+\vert E\vert+\vert V\vert rl)$. In summary, the time complexity of the feature extraction module depends on the length of the input time series. By controlling the length of the input time series, the time complexity of each batch can be reduced. Furthermore, the time complexity of the Conditional Separation Module and the prediction head is similar to that of an MLP, which has a relatively low impact on the overall complexity of the model. This highlights the efficiency of SC-FSGL in handling computational demands, as it remains effective even when processing high-dimensional data. 

\subsection{Proof of Convergence Analysis}
\label{appendix:proof}
Suppose there are a total of $K$ clients, $\mathcal{G}_{k,t}$, representing the space-time graph input of the client $k(k\in K)$ at the edge of time $t$, and the local model parameter is defined as $\theta^k_r=\{{\theta^k_{r, a},\theta_{r,b}}\}$, where $r\in R$ represents the aggregation Round, $\theta_{r, a}$ represents a local model parameter that does not participate in aggregation, $\theta_{r,b}$ is a model parameter of causal codebook, and $\theta_{r,b}$ participates in global aggregation. Partial loss function is $L (\theta_r ^ k) = L (\theta_ {r, a} ^ k) + L (\theta ^ k_ {r, b})$. To obtain the convergence analysis of the model, we make the following assumptions about $L_k(\theta_r^k)$ :
 \begin{align}
  \label{eq:1}
  &\frac{1}{R}\sum_{r=1}^{R}[L_k(\theta^k_r)-L_k(\theta^{k,*})]\to 0,\\
  &\quad \theta^{k,*}\triangleq\ \arg\min_{\theta_r}\sum_{r=1}^{R}L_k(\theta^{k}_r)
\end{align}
When $L_k$ is convex, according to assumption~\ref{assum.1}, we have
\begin{align}
  \label{eq:2}
L_k(\theta^{k,*}) \geq L_k(\theta^k_r) + \langle \ \nabla L_k(\theta^k_r),\theta^{k,*}-\theta^k_r\rangle
\end{align}
Add up both sides with the rounds and take the mean
\begin{align}
  \label{eq:3}
\frac{1}{R}\sum_{r=1}^{R}[L_k(\theta^k_{r})-L_k(\theta^{k,*})]\leq \frac{1}{R}\sum_{r=1}^{R}\langle L_k(\theta^k_r),\theta^k_r-\theta^{k,*}\rangle
\end{align}
According to the gradient descent formula $\theta^k_{r+1}=\theta^k_r-\eta_t \nabla L_k(\theta^k_r)$, subtracting $\theta^{k,*}$ from both sides, we have
\begin{align}
  \label{eq:4}
&\| \theta^k_{r+1} - \theta^{k,*} \|_2^2 \nonumber\\
&= \| \theta^k_r - \theta^{k,*} - \eta_t\nabla L_k(\theta^k_r) \|_2^2 \nonumber \\
&= \| \theta^k_r - \theta^{k,*} \|_2^2 - 2\eta_t \langle \nabla L_k(\theta^k_r), \theta^k_r - \theta^{k,*} \rangle + \eta_t^2 \| \nabla L_k(\theta^k_r)\|_2^2
\end{align}
We have 
\begin{align}
  \label{eq:5}
&\langle \nabla L_k(\theta^k_r), \theta^k_r - \theta^{k,*} \rangle \nonumber\\
&= \frac{1}{2\eta_t} [ || \theta^k_r - \theta^{k,*} ||_2^2 - || \theta^k_{r+1} - \theta^{k,*} ||_2^2 ] + \frac{\eta_t}{2} || \nabla L_k(\theta^k_r)||_2^2
\end{align}
Bring equation.\ref{eq:5} into the equation.\ref{eq:3}
\begin{align}
  \label{eq:6}
&\frac{1}{R}\sum_{r=1}^{R}[L_k(\theta^k_r)-L_k(\theta^{k,*})] \nonumber\\
&\leq \frac{1}{R}\underbrace{ {\sum_{r=1}^R \frac{1}{2\eta_t}}[|| \theta^k_r-\theta^{k,*}||_2^2-||{\theta^k_{r+1}-\theta^{k,*}||_2^2}]}_{A}\nonumber\\
&+\frac{1}{R}\underbrace{\sum_{t=1}^T \frac{\eta_t}{2} ||\nabla L_k(\theta^k_r)||_2^2}_B
\end{align}
For B, If $L_k$ is a convex function, then $\theta$ is a convex set, and we assume a bound between the model parameters and the optimal model parameters, where $||\theta_r^k-\theta^{k,*}||\leq I$
\begin{align}
  \label{eq:7}
B \leq \sum_{r=1}^R \frac{\eta}{2} I^2 
\end{align}
For A, we expand and sum
\begin{align}
\label{eq:8}
A&= \frac{1}{2\eta_t} \| {\theta^k}_1 - {\theta}^{k,*} \|_2^2 - \frac{1}{2\eta_1} \| {\theta^k}_2 - {\theta}^{k,*} \|_2^2 \nonumber\\
&+ \frac{1}{2\eta_2} \| {\theta^k}_2 - {\theta}^{k,*} \|_2^2 - \frac{1}{2\eta_2} \| {\theta^k}_3 - {\theta}^{k,*} \|_2^2 \nonumber\\
&+ \cdots \quad\ \nonumber\\
&+ \frac{1}{2\eta_R} || {\theta^k}_R - {\theta}^{k,*} ||_2^2 - \frac{1}{2\eta_R} | |{\theta^k_{R+1}-\theta^{k,*}}||^2_2 \nonumber\\
&=\underbrace{\frac{1}{2\eta_1}||\theta^k_1-\theta^{k,*}||_2^2}_{A_1} \nonumber\\
&+\underbrace{\sum_{r=2}^R ( \frac{1}{2\eta_r}}_{A_2} - \underbrace{\frac{1}{2\eta_{r-1}} ) || \theta^k_r-\theta^{k,*}||-\frac{1}{2\eta_R}||\theta^k_{R+1}-\theta^{k,*}||}_{A_3}
\end{align}
According to Assumption~\ref{assum.2}
\begin{align}
  \label{eq:8_1}
  A_1 = \frac{1}{2\eta_1} \| {\theta^k}_1 - {\theta}^{k,*} \|_2^2 \leq \frac{1}{2\eta_1} I^2 
\end{align}
\begin{align}
  \label{eq:9}
  &A_2 = \sum_{r=2}^R ( \frac{1}{2\eta_r} - \frac{1}{2\eta_{r-1}} ) || {\theta^k}_r - {\theta}^{k,*} ||_2^2 \nonumber\\
  &\leq I^2 \sum_{r=2}^R ( \frac{1}{2\eta_r} - \frac{1}{2\eta_{r-1}} )
\end{align}
\begin{align}
  \label{eq:10}
  -A_3 = -\frac{1}{2\eta_R} \| {\theta^k}_{R+1} - {\theta}^{k,*} \|_2^2 \leq 0 
\end{align}
Substitute the equations~\ref{eq:8},\ref{eq:9}, and~\ref{eq:10} into A
\begin{align}
  \label{eq:11}
  A \leq \frac{1}{2\eta_1} I^2 + I^2 \sum_{r=2}^R ( \frac{1}{2\eta_t} - \frac{1}{2\eta_{t-1}} ) + 0 = I^2 \frac{1}{2\eta_R}
\end{align}
Substitute~\ref{eq:7} and~\ref{eq:12} into~\ref{eq:6}
\begin{align}
  \label{eq:12}
\frac{1}{R}\sum_{r=1}^{R}[L_k(\theta^k_r)-L_k(\theta^{k,*})] \leq I^2 \frac{1}{2R\eta_R} + \frac{M^2}{2R} \sum_{r=1}^R \eta_r
\end{align}
Since $L (\theta_r ^ k) = L (\theta_ {r, a} ^ k) + L (\theta ^ k_ {r, b}) $
\begin{align}
  \label{eq:13}
\frac{1}{R}&\sum_{r=1}^{R}[L_k(\theta_r^k)-L_k(\theta^{k,*}) \nonumber\\
&= \frac{1}{R}\sum_{r=1}^R [L_k(\theta_{r,a}^k)+L_k(\theta^k_{r,b})-L_k(\theta_{r,a}^{k,*})-L_k(\theta^{k,*}_{r,b})] \nonumber\\
&=\frac{1}{R}[L_k(\theta^k_{r,a})-L_k(\theta_a^{k.*}))]\nonumber\\ 
&+ \frac{1}{R}\sum_{r=1}^{R}[L_k(\theta^k_{r,b})-L_k(\theta_{b}^{k,*})]
\end{align}
For causal codebook
\begin{align}
  \label{eq:14}
\frac{1}{R}&\sum_{r=1}^{R}[L_k(\theta^k_{r,b})-L_k(\theta_{b}^{k,*})] \nonumber\\
&=\frac{1}{R}\sum_{r=1}^{R}[L_k(\theta_r^k)-L_k(\theta^{k,*})- \frac{1}{R}[L_k(\theta^k_{r,a})-L_k(\theta_a^{k.*}))]
\end{align}
Because $L (\theta ^ k_ {r, a})$ satisfy the properties of convex function, so the $L (\theta ^ k_ {r, a}) - L (\theta_a ^ {k, *})) \geq 0$, then
\begin{align}
  \label{eq:15}
\frac{1}{R}&\sum_{r=1}^{R}[L_k(\theta^k_{r,b})-L_k(\theta_{b}^{k,*})]\nonumber\\
&\leq \frac{1}{R}\sum_{r=1}^{R}[L_k(\theta_r)-L_k(\theta^{k,*})] \nonumber\\
&\leq I^2 \frac{1}{2R\eta_R} + \frac{M^2}{2R}\sum_{r=1}^R \eta_r
\end{align}
When the learning rate is fixed, $\eta_r=\eta$
\begin{align}
  \label{eq:16}
\frac{1}{R}&\sum_{r=1}^{R}[L_k(\theta^k_{r,b})-L_k(\theta_{b}^{k,*})] \nonumber\\
&\leq \frac{1}{R}\sum_{r=1}^{R}[L_k(\theta_r)-L_k(\theta^{k,*})] \nonumber\\
&\leq I^2 \frac{1}{2R\eta} + \frac{M^2}{2}\eta
\end{align}
When leaning rate $\eta_r = \frac {1} {r ^ {\frac {1} {2}}}$, bring it into equation~\ref{eq:14}
\begin{align}
  \label{eq:17}
\frac{1}{R}&\sum_{r=1}^{R}[L_k(\theta^k_{r,b})-L_k(\theta_{b}^{k,*})] \nonumber\\
&\leq \frac{1}{R}\sum_{r=1}^{R}[L_k(\theta_r)-L_k(\theta^{k,*})] \nonumber\\
&\leq I^2 \frac{R^{\frac{1}{2}}}{2R\eta_R} + \frac{M^2}{2R}\sum_{r=1}^R \frac{1}{r^{\frac{1}{2}}} \nonumber\\
&\leq I^2 \frac{R^{\frac{1}{2}}}{2R\eta_R} + \frac{M^2}{2R}(1+\int_{1}^{R} \frac{dr}{r^{1/2}}) \nonumber\\
&=(I^2+M^2) \frac{1}{2R^{\frac{1}{2}}} -\frac{1}{1R}M^2
\end{align}

\end{document}